\PassOptionsToPackage{table}{xcolor}
\documentclass[]{xiaomiev}

\usepackage{amsmath,amssymb}
\usepackage{array}
\usepackage{enumitem}
\usepackage{float}
\usepackage{longtable}
\usepackage{makecell}
\usepackage{needspace}
\usepackage{pifont}
\usepackage{siunitx}
\usepackage{tabularx}

\usepackage{booktabs}
\usepackage{multirow}
\usepackage[table]{xcolor}
\usepackage{graphicx}

\microtypesetup{expansion=false}

\usepackage{pifont}
\newcommand{\xmark}{\ding{55}}
\newcommand{\cmark}{\ding{51}}
\usepackage{xspace}
\newcommand{\teacher}{DriveRL\xspace}
\newcommand{\student}{DriveZero\xspace}

\newcommand{\cmtt}[1]{{\fontfamily{cmtt}\selectfont #1}}

\definecolor{HermesOrange}{HTML}{FF7E00}
\definecolor{HermesTint}{HTML}{FFF8F2}
\definecolor{HermesLine}{HTML}{E6D8CC}
\definecolor{PlaceholderFill}{HTML}{F5F5F5}
\definecolor{PlaceholderLine}{HTML}{A8A8A8}
\definecolor{modelclr}{RGB}{226,110,92}  
\definecolor{humanclr}{RGB}{86,163,74}   
\definecolor{drivorclr}{RGB}{230,141,73} 
\hypersetup{
  pdftitle={DriveZero: End-to-End Driving Beyond Human Demonstrations},
  pdfauthor={Xiaomi EV},
  linkcolor=HermesOrange,
  citecolor=HermesOrange,
  urlcolor=HermesOrange
}

\setlist[itemize]{leftmargin=15pt,itemsep=2pt,topsep=3pt}
\setlist[enumerate]{leftmargin=18pt,itemsep=2pt,topsep=3pt}
\renewcommand{\arraystretch}{1.08}

\makeatletter
\providecommand{\@bottomtitlebar}{}
\makeatother

\title{DriveZero: End-to-End Driving Beyond Human Demonstrations}

\author{{\bfseries AD \& Robotics, L3 Team}\\[1.5mm]
        {\normalsize\mdseries Xiaomi EV}}
\checkdata[Report]{Technical Report, September 2026}
\checkdata[Website]{\url{https://xiaomiautol3.github.io/DriveZero}}
\abstract{
Most end-to-end autonomous-driving systems learn by imitating human driving logs, leaving their learned behavior constrained by the quality and behavioral coverage of the recorded trajectories. This report presents DriveZero, an end-to-end system that learns driving behavior beyond human demonstrations. It decomposes driving into a perception model and an action model, pretrains each in the regime best suited to it, and combines them into one end-to-end planner. The two models call for different learning recipes: perception must understand the world, and benefits from massive and diverse visual data; action must interact with it, and requires closed-loop feedback. On the action side, we introduce \teacher, a mixed-agent closed-loop reinforcement-learning framework. It converts real driving logs into interactive worlds, where a privileged teacher policy is trained with PPO through closed-loop rollouts. For the perception model, DriveVFM consolidates multiple frozen vision foundation models, including DINOv3, SigLIP2, SAM and Depth Anything V2, into a single backbone from raw images alone, requiring no task-specific annotations. \student then unifies the two: a camera-only planner that distills the frozen \teacher teacher through its rolled-out trajectories. The goal-conditioned teacher can moreover be queried under augmented driving intents, yielding diverse, goal-consistent supervision that logged data cannot provide. On nuPlan, \teacher with value-guided test-time action search achieves a mean score of 93.57 across the Val14, Test14-hard, and Test14-random community splits in both non-reactive and reactive modes, exceeding the Log-Replay expert on all three splits. \student achieves state-of-the-art performance on NAVSIMv1, NAVSIMv2 and the
closed-loop HUGSIM benchmark without any human trajectory supervision: it reaches 95.3 PDMS on navtest, surpassing the human driver (94.8), 57.1 EPDMS on navhard, and 46.6 HD-Score zero-shot on HUGSIM.
}

\begin{document}
\maketitle

\section{Introduction}
\label{sec:introduction}
End-to-end autonomous driving aims to map onboard observations and navigation intent directly to vehicle motion~\cite{chen2023e2esurvey}. Most existing systems learn this mapping by imitating human driving logs~\cite{cheng2024plantf,cheng2024pluto,hu2023uniad,jiang2023vad,jia2025drivetransformer,li2025hydramdp,liao2025diffusiondrive}. This paradigm is scalable and stable, but it makes the recorded human trajectory the sole source of supervision, leaving the learned behavior limited by the quality and coverage of the logs. Each logged scene contains only one realized future, even when multiple actions would be valid; safety-critical deviations and recovery maneuvers are rare; and states induced by the learned policy are absent from the offline data. Consequently, errors can compound once the policy leaves the demonstration distribution~\cite{ross2011dagger}. 

\begin{figure*}[t]
  \centering
  \includegraphics[width=\textwidth]{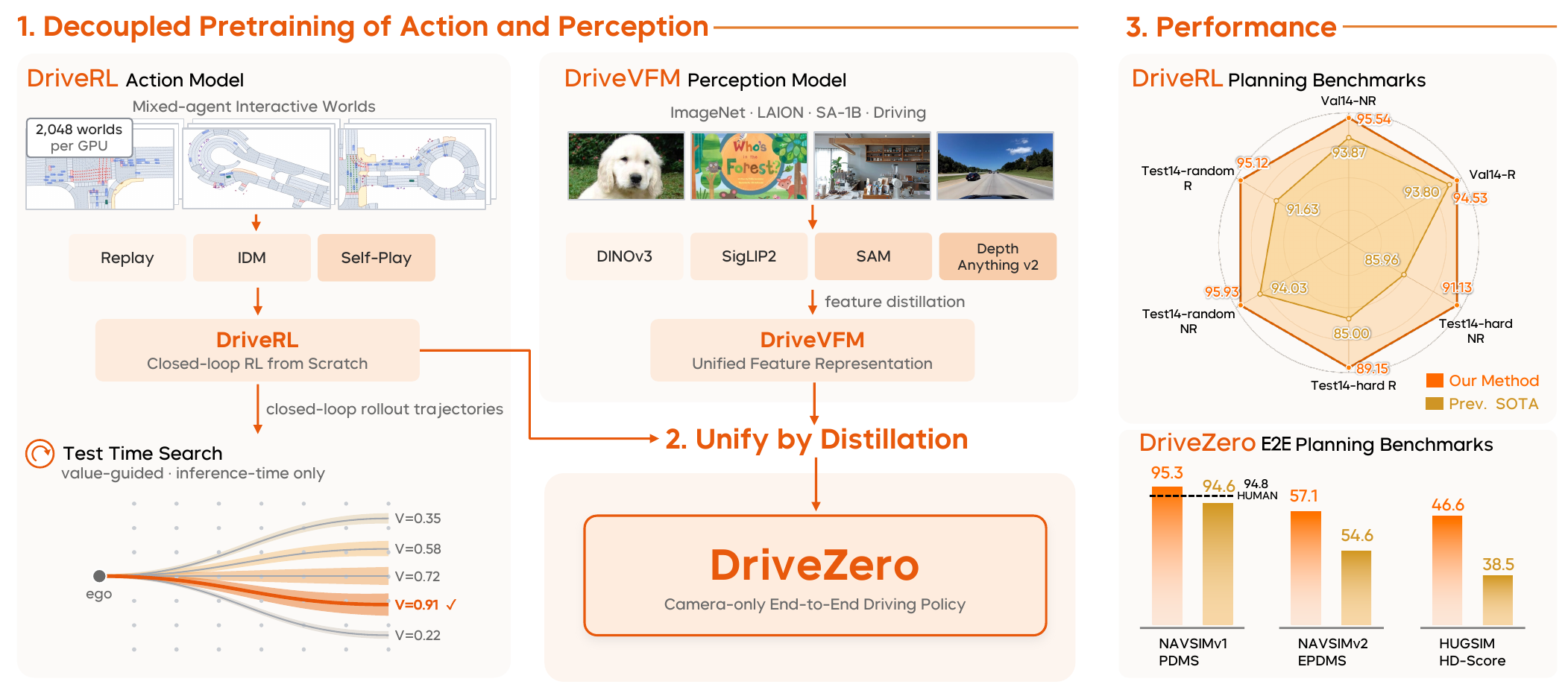}
  \caption{\textbf{Overview of our end-to-end driving system.} \textbf{(1) Decoupled pretraining.} The action model, \teacher, is a privileged policy trained from scratch with closed-loop RL in mixed-agent interactive worlds built from real nuPlan logs. At inference, a value-guided test-time search rolls out several sampled actions, scores them using short-horizon rewards and the critic, and conditionally replaces the modal action with a higher-value candidate. The perception model, DriveVFM, distills frozen vision foundation models into a single driving backbone from raw images, without task-specific labels. \textbf{(2) Unification by distillation.} \student initializes its encoder from DriveVFM and learns from \teacher rollouts instead of human trajectories, yielding a camera-only end-to-end planner. \textbf{(3) Performance.} \teacher exceeds the log-replay expert and prior methods on all six nuPlan closed-loop settings; \student surpasses the human driver on NAVSIMv1 and sets the state of the art on NAVSIMv2 and HUGSIM without any human trajectory supervision.}
  \label{fig:overall-framework}
\end{figure*}

Reinforcement Learning (RL) offers a different source of driving behavior. By optimizing explicit objectives through closed-loop interaction, an RL policy can observe the consequences of its own actions, turn failures into training signal, visit perturbed states, and learn how to recover from them~\cite{scheel2021urbandriver,cusumano2025selfplay,jaeger2025carl}. Thus, RL is not restricted to reproducing the single action sequence chosen by a human driver; it can optimize behavior beyond the demonstrations contained in the logs. Bringing this advantage to end-to-end driving, however, is difficult: direct visual RL couples sample-intensive exploration with costly visual simulation, limiting its scalability~\cite{rowe2026gigapixel}. This creates a central tension: closed-loop RL is well suited to learning how to drive, whereas the camera-only policy required for deployment is poorly suited to large-scale online exploration. Privileged RL teachers followed by visual-policy distillation \citep{zhang2021roach} have recently regained attention as a way around this bottleneck \citep{rowe2026gigapixel,xiong2026terratransfer}. ROACH distills a CARLA RL coach into a monocular policy, while Gigapixel transfers a vector-observation RL teacher to a pixel-based student through self-play DAgger. These methods train the policy in closed loop but take the visual encoder off the shelf. We instead pretrain a perception model and an action model separately, each in the regime best suited to it: perception benefits from large, diverse image collections and rich representation supervision, whereas action benefits from high-throughput closed-loop interaction over compact structured states. We then reunify the two through distillation into a deployable camera-only policy.

For the action model, we introduce \textbf{\teacher}, a mixed-agent closed-loop reinforcement-learning framework. \teacher converts real nuPlan logs \citep{caesar2021nuplan} into interactive worlds, where each traffic participant receives an independent behavior provider through a common physical state–action interface, so that log replay, rule-based behaviors, and learned policies coexist within one scene. In these worlds we train the teacher of our system, a privileged policy, with PPO \citep{schulman2017ppo} through closed-loop rollouts.
The resulting 5.70M-parameter teacher policy observes structured scene state and navigation, outputs bounded Beta distributions over longitudinal jerk and tire steering-angle rate, and controls the vehicle directly without trajectory refinement. PPO also trains a value function (a.k.a critic). We exploit it by a value-guided action search in test-time to cope with out of domain cases for the learned policy. It samples several top actions from the policy, rollouts for a few steps. The final policy is replaced by the best sampled action only if its estimated return exceeds that of the policy mode by a fixed margin. On nuPlan, \teacher achieves a mean score of 93.01 across the Val14, Test14-hard, and Test14-random \citep{dauner2023pdmclosed,cheng2024plantf} community splits, exceeding the Log-Replay expert in all three sets, and test-time search raises the mean further to 93.57. The learned behavior thus already surpasses the demonstrations that seed its training worlds.

We build the perception model with \textbf{DriveVFM}. Rather than coupling a backbone with multiple annotated perception tasks, DriveVFM consolidates frozen vision foundation models (DINOv3 \citep{simeoni2025dinov3}, SigLIP2 \citep{tschannen2025siglip2}, SAM \citep{kirillov2023sam}, and Depth Anything V2 \citep{yang2024depthanythingv2}) into a single driving backbone from raw images alone, following the agglomerative distillation of RADIO \citep{ranzinger2024amradio}. Since the supervision comes entirely from frozen foundation-model features, training requires no detection, segmentation, lane, or depth labels. This allows the pretraining corpus to freely mix web-scale imagery \citep{schuhmann2022laion5b, kirillov2023sam, russakovsky2015imagenet} with driving scenes \citep{yang2024genad,caesar2021nuplan,waymo}, so the backbone acquires driving-relevant semantics, geometry, and spatial structure at scale. On top of this backbone, \textbf{\student} completes the system: a camera-only planner that encodes multi-view images with DriveVFM, decodes multiple trajectory proposals, and learns through winner-takes-all distillation against trajectories rolled out by the frozen \teacher teacher, with a learned scoring head ranking the proposals at inference. Its training signal thus comes from reinforcement-learned behavior rather than from human demonstrations. Because the teacher is goal-conditioned, the same scene can further be queried under augmented driving intents, yielding diverse, goal-consistent supervision that logged data cannot provide. This augmentation significantly diversifies the training data, increasing the available supervision signals in the training.

\student reaches 94.8 PDMS on NAVSIMv1 navtest without any human trajectory supervision, on par with the human driver (94.8). Scaling its training data with simulation~\cite{tian2026simscale} sets the state-of-the-art on all three benchmarks, with 95.3 PDMS on navtest, 57.1 EPDMS on NAVSIMv2 navhard, and 46.6 HD-Score on the closed-loop HUGSIM benchmark. In ablation studies, teacher supervision with goal augmentation surpasses human-trajectory supervision, and an additive study shows that each teacher foundation model contributes cumulatively to DriveVFM. Together, these results validate the decomposition: a reinforcement-learned action model freed from human demonstrations, a perception model pretrained without task annotations, and a distillation that unifies them into a camera-only planner surpassing imitation-based counterparts.


\section{Methodology}
\label{sec:methods}
Our methodology consists of three stages: learning driving behavior, pretraining visual representations, and transferring the learned behavior to a camera-only policy. Section \ref{sec:teacher} describes \teacher, including its structured inputs and policy network, mixed-agent simulation, and RL objective. Its learned critic further supports value-guided test-time action search. Section \ref{sec:drivevfm} introduces DriveVFM and its multi-teacher visual representation distillation. Finally, Section \ref{sec:drivezerostudent} presents \student, which reunifies the two pretrained models through multimodal trajectory distillation, proposal scoring, and goal-conditioned augmentation.
\subsection{\teacher: Learning a Privileged Teacher from Scratch with Closed-Loop RL}
\label{sec:teacher}
\teacher is a closed-loop reinforcement learning system that trains a privileged teacher driving policy from scratch, following the recent evidence that large-scale closed-loop RL alone can produce robust driving behavior \cite{cusumano2025selfplay,jaeger2025carl}. At each step, it receives a privileged structured observation together with goal points that express navigation intent and outputs a distribution over actions that control longitudinal jerk and tire steering-angle rate. Closed-loop training requires an interactive world in which the ego vehicle rolls out its own actions and the surrounding traffic participants evolve alongside it, either by replaying their real driving behavior or by following dynamically consistent models that react to the ego. \teacher builds such worlds from real nuPlan logs using a mixed-agent simulator. Each background actor follows log replay, a rule-based model, or a learned policy, while the simulator runs up to 196,608 worlds in parallel across 96 GPUs. Within these worlds, \teacher is optimized with PPO \cite{schulman2017ppo} against a reward that combines hard safety events, goal arrival, and soft driving-quality terms. It starts from random initialization and receives no imitation pretraining; logged data only seeds the scenes and navigation goals.

\subsubsection{The Structured Inputs and Policy Network}
\label{sec:structured-teacher}
At each step $t$, the structured observation $O_t$ contains the ego vehicle, surrounding traffic participants, the local vector map, and traffic-light states.
The ego and each traffic participant are represented by five frames in total, including the current frame and four preceding frames sampled at 5 Hz, up to 96 participants; the map is represented by up to 256 tokens of local vector elements, with traffic-light states attached to the elements they govern. Navigation information is provided by goal points $G_t$ in the ego frame.

\paragraph{Goal construction.}
The goal points specify the positions the ego vehicle should reach in the near future. 
During training, \teacher constructs a two-point goal representation from future ego positions in the log, using either the same position for both points or two distinct positions. 
At deployment, it selects two goal points, a near and a far anchor, along the current route. Their look-ahead distances scale with the vehicle's speed, and the anchors are recomputed at every step. Despite their different construction, both training and deployment goals use the same two-point, permutation-invariant representation.

\paragraph{Policy network.}
Three dedicated encoders embed the agent, ego-kinematics, and map inputs to 256-dimensional tokens. The two ego-frame goal points are embedded using sinusoidal positional encoding \citep{vaswani2017attention} followed by a shared MLP, then mean-pooled into a single goal feature. Using the ego token as the only query, two cross-attention layers aggregate agent context, followed by an ego-to-map cross-attention layer over map tokens to extract map context. The resulting ego token, goal and kinematics embeddings, and map context are concatenated and fused by a shared MLP before entering the policy head. The complete policy contains 5.7M parameters.

\paragraph{Beta action distribution.}
The action head defines independent Beta distributions for normalized jerk and steering-rate commands \citep{chou2017beta}. Following CaRL \citep{jaeger2025carl}, we restrict the shape parameters to $(1, \infty)$ by adding 1 to the softplus activation function, which keeps each distribution unimodal and avoids overweighting extreme commands.
PPO samples actions during training, while evaluation uses the mode of each distribution. The normalized commands are then mapped to bounded physical commands and executed by a kinematic bicycle model (Appendix~\ref{app:policy-dynamics}).

\subsubsection{Mixed-Agent Simulation}
\label{sec:mixed-agent-worlds}

\paragraph{Agent behavior composition.}
The simulator assigns each background actor an independent behavior provider through a common physical state--action interface. The runtime supports three provider categories: log replay, rule-based behaviors such as IDM and front-vehicle braking, and learned policies such as Self-Play. This allows heterogeneous behaviors to coexist in one scene. In the reported training setup, background vehicles use a batched approximation of IDM, a small subset may be controlled by front-vehicle braking, and pedestrians and other non-vehicle actors remain on log replay. The Self-Play provider is described next.

\paragraph{Self-play extension.}
Inspired by GigaFlow~\citep{cusumano2025selfplay}, the mixed-agent runtime also supports a learned provider. The policy under training can control selected background vehicles in addition to the ego. Each policy-controlled vehicle receives its own ego-centric observation and goal, and their actions jointly advance the scene, while the remaining actors stay under IDM or log replay. By default, \teacher controls only the ego vehicle. Self-play is the configuration we use for real-world deployment (Section \ref{sec:realworld}). Rule-based or replayed traffic does not respond to the ego the way real drivers do, whereas policy-controlled background vehicles expose \teacher to more natural interactions during training. On nuPlan the benefit is small. We attribute this to the benchmark itself, whose closed-loop evaluation drives background vehicles by IDM and therefore does not reward more natural interaction behavior. Tables~\ref{tab:appendix-selfplay} and~\ref{tab:self-play-comparison} report the configuration and results.

\paragraph{Scene-consistent actor insertion.}
During closed-loop RL training, \teacher initializes each world from a recorded nuPlan scene and background actors enter the scene according to the log. Some actors appear only later in the recording. Inserting them at the logged entry time is consistent with the original log, but once the ego has deviated from the logged trajectory during rollout, an inserted actor may land in an implausible position relative to the ego, or even collide with it. \teacher therefore gates the insertion with a scene-consistency check: a late-entering actor is released only when it is present in the current log frame, the ego pose is within 5 m and 0.35 rad of the logged ego pose, and it would not collide with any visible actor. Once released, the actor follows its logged trajectory. 

\paragraph{Parallel PPO training.}
The simulator implements the entire rollout loop as batched GPU operations, including background-agent behaviors, vehicle dynamics, reward computation, and scene editing. \teacher uses 2,048 worlds per rank. We also find that a larger PPO minibatch improves training performance. Rollout and optimization settings are specified in Section~\ref{sec:driveRL-results} and Appendix~\ref{app:simulation-training}.

\paragraph{Training objective.}
\teacher is optimized using hard-event penalties $h_t$, a one-time goal-arrival bonus $g_t$, and six soft driving-quality scores $q_{t,k}\in[0,1]$ covering safety, compliance, and comfort. Hard events terminate the rollout, whereas goal arrival does not. Let $d_t$ indicate a hard termination and $\mathcal{K}$ denote the six soft criteria. The scalar reward is
\begin{equation}
  r_t
  =
  h_t
  +
  (1-d_t)
  \left(
    g_t
    +
    \frac{1}{H}
    \prod_{k\in\mathcal{K}} q_{t,k}
  \right),
  \label{eq:teacher-reward}
\end{equation}
where $H=110$ is the rollout horizon. The multiplicative term requires the soft criteria to be jointly satisfied. PPO optimizes this scalar reward. The critic is decomposed into one value channel per reward term, with the channels summing to the total return. Full reward and critic details are given in Appendix~\ref{app:reward-ppo}.

\subsubsection{Value-Guided Test-Time Action Search}
\label{sec:tts-method}

PPO compresses the behavior discovered during closed-loop training into an action distribution, but modal execution discards both the policy's local action diversity and the critic's estimate of delayed return. We use these signals for value-guided action reranking at inference time, following the broader principle of using policy and value estimates to guide test-time search (TTS)~\cite{silver2017mastering,schrittwieser2020mastering}. Rather than training a separate planner, the procedure compares alternative first actions using the same ego dynamics, reward definition, and discount factor as PPO, together with a short-horizon approximation of background-actor motion.

\paragraph{Policy-supported action proposals.}
At each planner update, we retain the mode of the Beta policy as candidate $0$ and sample $N-1$ alternative first actions from the same policy. Retaining the modal action makes the search a conservative extension of the deployed policy. Additional computation changes the decision only when a policy-supported alternative is predicted to be meaningfully better.

\paragraph{Short-horizon closed-loop evaluation.}
Each candidate is evaluated through a rollout of $L$ transitions. The candidate determines the first ego action, while subsequent actions are given by the mode of the same policy from the candidate-specific states. Candidates therefore differ not only in their initial control but also in the continuation induced by the resulting state. During rollout, background actors are extrapolated using a constant-turn-rate-and-acceleration model with bounded acceleration and yaw rate. The rollout is deliberately short, intended to expose immediate differences in safety, route compliance, and comfort rather than to model long-horizon traffic interaction.

\paragraph{Reward-consistent scoring.}
For candidate $i$, let $r_\ell$, $d_\ell$, and $O_\ell$ denote the reward, termination indicator, and observation at TTS rollout step $\ell$, where $r_\ell$ and $d_\ell$ are computed online along the rollout.
The candidate is scored by
\begin{equation}
    S^{(i)}
    =
    \sum_{\ell=0}^{L-1}
        \gamma^{\ell}
        \prod_{u<\ell} \bigl(1 - d_u\bigr)\,
        r_\ell
    +
    \gamma^{L}
    \prod_{u<L} \bigl(1 - d_u\bigr)\,
    V\!\left(O_L\right),
    \label{eq:tts-score}
\end{equation}
where $\gamma$ is the discount factor, $V$ is the teacher critic's estimate of the discounted return.
A terminal transition thus contributes its own reward, while all later rewards and the bootstrap are masked.
Reusing the training reward, discount factor, and critic keeps candidate ranking aligned with the objective optimized by PPO.

\paragraph{Conservative action selection.}
Because both the background-actor rollout and the critic are approximate, a small score difference may be noise rather than a better action. We therefore execute the highest-scoring sampled candidate only if its score exceeds that of the modal action by a margin $\delta$, and execute the modal action otherwise.

\paragraph{Computation.}
All candidates are rolled out as a single batch, so increasing the number of candidates $N$ primarily enlarges parallel computation, whereas increasing the TTS horizon $L$ adds sequential policy, dynamics, and reward evaluations. The two parameters therefore provide distinct ways to exchange inference-time computation for action-selection quality without retraining or modifying the policy.

\subsection{DriveVFM: Distilling Multiple Vision Foundation Models into One Driving Backbone}
\label{sec:drivevfm}

The teacher policy learns from privileged structured observations, whereas the deployable student must infer the information needed for planning directly from camera images. Transferring the driving capability of the teacher policy therefore first requires a strong visual backbone that recovers driving-relevant semantics, geometry, and spatial structure. Driving-oriented representation learning conventionally trains a shared backbone with multiple auxiliary perception heads, such as detection, lane estimation, segmentation, and depth \citep{teichmann2018multinet,wu2022yolop,hu2023uniad}, which depends on large quantities of task-specific annotations. General purpose vision foundation models offer transferable representations without any of this, but no single model captures all driving-relevant capabilities equally well.

\begin{figure*}[t]
  \centering
  \includegraphics[width=0.95\textwidth]{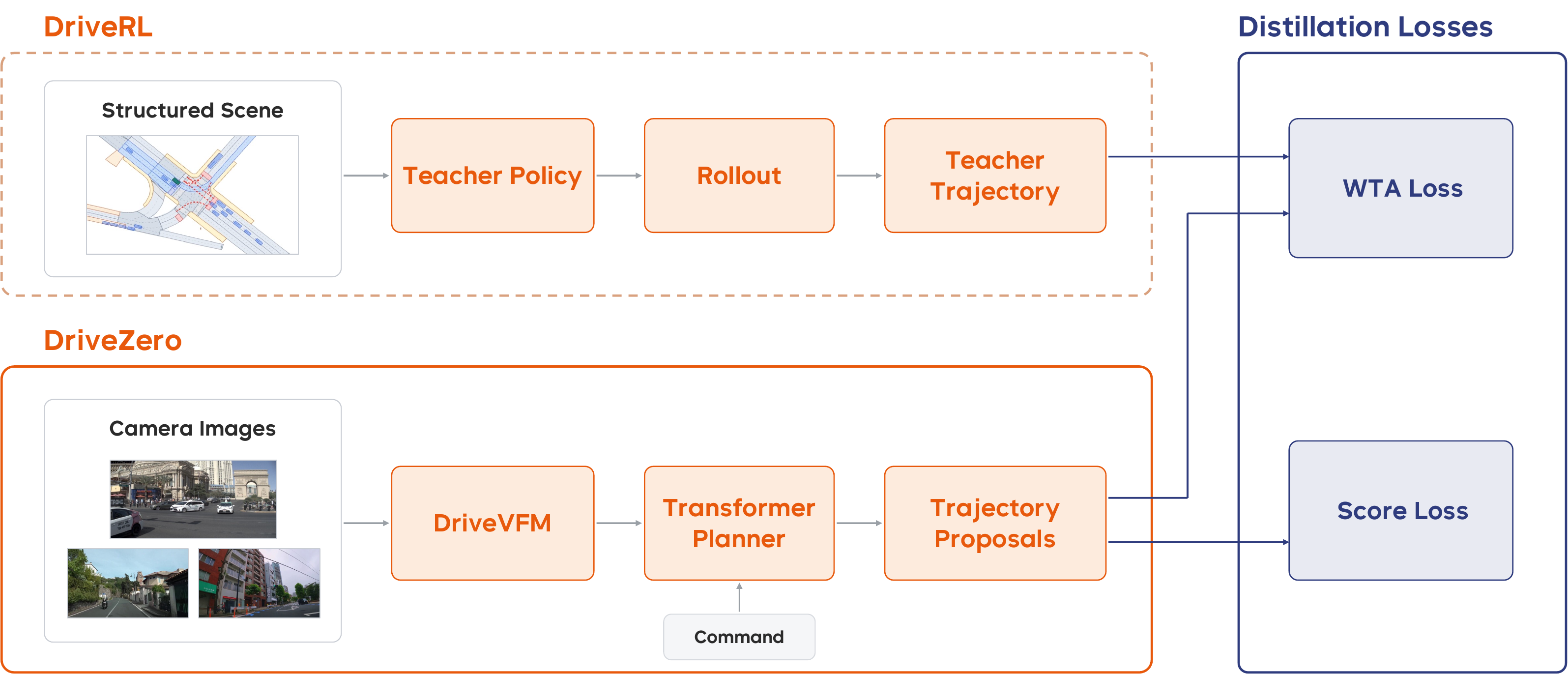}
  \vspace{3pt}
  \caption{\textbf{Overview of the end-to-end student policy distillation framework.} The privileged \teacher teacher rolls out a 20-step trajectory from structured scene observations. The camera-only student \student uses DriveVFM features and a command-conditioned Transformer planner to generate multiple 20-step trajectory proposals. Winner-takes-all (WTA) trajectory supervision transfers the teacher behavior, while the proposal-scoring branch is trained against PDM targets. The teacher is used only during training and is removed at inference time.}
  \label{fig:ad-distillation-framework}
\end{figure*}

DriveVFM replaces manually annotated auxiliary tasks with heterogeneous frozen foundation models, each acting as a learned proxy for a family of perception objectives: DINOv3 \citep{simeoni2025dinov3} provides spatial structure and visual correspondence, SigLIP2 \citep{tschannen2025siglip2} contributes image-level semantics and open-vocabulary information, SAM \citep{kirillov2023sam} supplies boundary-sensitive features analogous to segmentation supervision, and Depth Anything V2 \citep{yang2024depthanythingv2} offers geometric cues. By directly matching their frozen features, DriveVFM consolidates these complementary capabilities into a single backbone, following the agglomerative distillation of RADIO \citep{ranzinger2024amradio}.

\paragraph{Architecture and objective.} Given an image $I$, a shared Vision Transformer \citep{dosovitskiy2021vit} produces one summary token per foundation model and a common set of spatial patch tokens. The model-specific summary tokens let heterogeneous image-level semantics coexist, whereas the shared patch tokens form a unified dense representation for downstream spatial reasoning. Each frozen foundation model is associated with a lightweight adaptor containing separate projections for its summary and patch representations. 
The supervision is asymmetric, matching each model where its representation is strongest: DINOv3 supervises both summary and patch tokens; SigLIP2 supervises only the summary token; and SAM supervises only the patch tokens. As for Depth Anything, it can enter through the same interface by matching its visual features or distilling its depth predictions. We choose to distill its visual features rather than its depth predictions, as feature distillation performs better in our experiments. Summary features are matched by cosine distance, while spatial patch features are optimized by mean-squared error. For the set of foundation models $\mathcal{T}$, the objective is
\begin{equation}
    \mathcal{L}_{\mathrm{DriveVFM}}
    =\sum_{k\in\mathcal{T}}
    \left(
        \lambda_k^{\mathrm{sum}}\mathcal{L}_k^{\mathrm{sum}}
        +
        \lambda_k^{\mathrm{patch}}\mathcal{L}_k^{\mathrm{patch}}
    \right).
    \label{eq:drivevfm-objective}
\end{equation}

\paragraph{Balancing heterogeneous foundation models.}
These frozen models produce features with substantially different scales, variances, and anisotropic activation distributions. A model with larger or more concentrated activations may dominate the gradients and bias the shared DriveVFM backbone toward its feature space. Per-model loss weights cannot fix this, because the imbalance also exists across feature dimensions within a single model. DriveVFM therefore applies PHI Standardization (PHI-S)~\citep{ranzinger2024phis} to each model's patch features. With mean and covariance estimated offline, PHI-S centers the features, applies a PCA--Hadamard rotation that distributes variance equally across the transformed dimensions, and rescales them by a shared scalar. Every model thus presents a similarly conditioned regression target, so its activation statistics no longer determine its effective importance in the joint objective.
Summary features are matched by cosine distance and are already insensitive to magnitude.

\paragraph{Stabilizing backbone optimization.}
During training, query--key dot products in some attention layers can grow rapidly, producing excessive pre-softmax attention logits and eventually causing loss spikes or divergence. Query--Key Normalization \citep{henry2020query,dehghani2023scaling} can constrain this behavior by modifying the attention computation. DriveVFM instead adopts a simpler intervention inspired by QK-Clip from Kimi K2~\citep{moonshotai2025kimi}.

\paragraph{Pretraining schedule and downstream interface.}
DriveVFM uses a two-stage progressive-resolution schedule. The first stage operates at $256\times256$ resolution for 600K optimization iterations with a global batch size of 2,048 images. The second stage increases the resolution to $512\times512$ and continues for another 200K iterations at the same global batch size. Both stages use AdamW \citep{loshchilov2017adamw}. The training corpus combines general-purpose data from LAION-2B \citep{schuhmann2022laion5b}, ImageNet-21K \citep{russakovsky2015imagenet}, and SA-1B \citep{kirillov2023sam} with driving images sampled from OpenDV \citep{yang2024genad}, nuPlan \citep{caesar2021nuplan}, and Waymo \citep{waymo}. This mixture preserves broad visual coverage while exposing the backbone to road geometry, traffic participants, and visual conditions encountered in autonomous driving.

After pretraining, the foundation models and their adaptors are discarded. The DriveVFM backbone encodes each camera with shared weights and provides the visual tokens that \student consumes, serving as the camera-side counterpart of the privileged scene inputs used by the teacher policy.

\subsection{\student: Distilling the RL Teacher into a Camera-Only Planner}
\label{sec:drivezerostudent}
\label{sec:student-distillation}

\teacher acts on structured scene state, while an end-to-end planner must act on camera images. The driving log bridges these two observation spaces: each frame in the log pairs the multi-view images with the structured state, so a trajectory produced by the teacher from the state can supervise the student on the images. \student builds on this correspondence. It is a camera-only end-to-end planner trained open-loop on logged frames, and the training targets come from rolling out the frozen teacher at each frame. The student never enters an interactive environment and no rendering is required. It inherits the behavior that the RL teacher learned in closed loop, and no human demonstrations are used in its training. Figure~\ref{fig:ad-distillation-framework} illustrates the overall framework.

\paragraph{Architecture.}
\student takes the multi-view images of the current frame, the ego kinematics, and a navigation command as inputs. The navigation command is derived from the goal point given to the teacher~\citep{openscene2023}, so student and teacher share one driving intent. From these inputs, \student produces $M$ candidate trajectories, each with a predicted score. DriveVFM, fine-tuned with LoRA \citep{hu2021lora}, encodes each camera image, and the output visual tokens are enriched with 3D position embeddings \citep{liu2022petr}. Following DrivoR \citep{kirby2026drivor}, learnable registers \citep{darcet2024vision} then compress these tokens into a small set of scene tokens. The ego kinematics and the command are embedded into one ego token, which is added to $M$ learnable trajectory queries. A trajectory decoder maps these queries to the output trajectories through cross-attention to the scene tokens. A separate scoring decoder then encodes each candidate trajectory as a score query, attends to the visual tokens, and predicts the six components of the PDM score \citep{dauner2023pdmclosed}. Trajectory generation and proposal scoring are trained with separate objectives.

\paragraph{Teacher rollouts as supervision.}
For each logged frame, the frozen teacher receives the structured state and a goal point, and is then rolled out with all background actors following the driving log. The resulting teacher trajectory $\tau^T$ is represented for the student as a sequence of ego-relative $(x,y,\mathrm{yaw})$ poses and serves as the regression target of a winner-takes-all objective~\citep{guzmanrivera2012multiple}. Each candidate proposal output by the trajectory decoder is compared with $\tau^T$, and only the closest proposal contributes to the trajectory loss:
\begin{equation}
    \mathcal{L}_{\mathrm{traj}}
    =\min_m \operatorname{dist}\!\left(\hat{\tau}_m,\tau^T\right),
    \label{eq:drivezero-trajectory}
\end{equation}
where $\hat{\tau}_m$ is the $m$-th proposal and $\operatorname{dist}(\cdot,\cdot)$ is the $L_1$ distance averaged over points. This assignment allows different trajectory queries to capture distinct driving modes, leaving the selection among them to the proposal scoring branch.

\paragraph{Proposal scoring.}
For each proposal, the scoring decoder predicts six driving-quality components covering collision avoidance, drivable-area
compliance, progress, time to collision, comfort, and driving-direction compliance. 
These components are supervised by their corresponding PDM targets~\citep{dauner2023pdmclosed}. Let $\hat{s}_{mk}$ and $s_{mk}$ denote the predicted logit and target for component $k$ of proposal $m$. The scoring loss is
\begin{equation}
    \mathcal{L}_{\mathrm{score}}
    =\frac{1}{M}\sum_{m=1}^{M}\sum_{k=1}^{6}
    \operatorname{BCE}\!\left(\sigma(\hat{s}_{mk}),s_{mk}\right).
    \label{eq:drivezero-score}
\end{equation}
The components are aggregated according to the benchmark scoring rule. At inference time, \student selects the candidate with the highest predicted aggregate score. Candidate trajectories are detached before entering the scoring branch, separating trajectory generation from proposal ranking. The complete objective combines winner-takes-all trajectory distillation with proposal scoring:
\begin{equation}
    \mathcal{L}_{\mathrm{\student}}
    =\lambda_{\mathrm{traj}}\mathcal{L}_{\mathrm{traj}}
    +\lambda_{\mathrm{score}}\mathcal{L}_{\mathrm{score}}.
    \label{eq:drivezero-objective}
\end{equation}

\paragraph{Goal augmentation.}
Human driving logs contain only one realized trajectory under one recorded route intent. \teacher, however, is goal-conditioned and can be queried with alternative driving intents at the same initial scene state. We exploit this property by augmenting the route intent and its associated navigation command before generating teacher supervision. \teacher consequently produces diverse yet goal-consistent trajectories beyond the single behavior recorded in the log. The same augmented route is used to evaluate student proposals, keeping the navigation command, teacher trajectory, and proposal scores mutually consistent. Goal augmentation therefore expands each logged scene into multiple valid supervision targets and directly exploits the generative capability of the closed-loop RL teacher.
\section{Experiments}
\label{sec:results}

We evaluate \teacher on nuPlan~\citep{caesar2021nuplan} and \student on NAVSIMv1~\citep{Dauner2024navsim}, NAVSIMv2~\citep{Cao2025navsimv2}, and HUGSIM~\citep{zhou2024hugsim}. \teacher is evaluated in closed loop using structured scene states, whereas \student takes camera images and follows the standard sensor-input protocol of each benchmark. DriveVFM is evaluated through the downstream \student planner rather than on a separate benchmark. The following sections describe the evaluation protocols, main results, and ablation studies.

\subsection{\teacher}
\label{sec:driveRL-results}
\paragraph{Training.}
Each training world is initialized from a recorded nuPlan scene, and PPO generates all training experience through interaction. We draw 922,703 scenes from the nuPlan trainval split, exported at 10~Hz with 41 history and 200 future frames, and the data loader takes every second frame, giving a 5~Hz training rate. \teacher is trained with 2,048 worlds per rank on 96 GPUs, using 110-step rollouts at 5~Hz over a 1:1 mixture of log-replay and IDM scenes. Each update runs four PPO epochs on the discounted return with $\gamma=0.99$, and training takes 2,400 updates and approximately 21 hours. Table~\ref{tab:teacher-config} summarizes the architecture and control configuration. Complete rollout, optimization, and input specifications are in Appendix~\ref{app:teacher-implementation}.

\begin{table}[H]
    \centering
    \caption{\textbf{\teacher training configuration.} Core architecture and control settings. Complete implementation details are provided in Appendix.}
    \label{tab:teacher-config}
    \small
    \setlength{\tabcolsep}{5pt}
    \begin{tabularx}{\linewidth}{p{0.40\linewidth}X}
        \toprule
        \textbf{Component} & \textbf{Teacher configuration} \\
        \midrule
        Token width / attention heads & 256 / 4 \\
        Ego-to-agent / ego-to-map attention & 2 / 1 layers \\
        Agent history & 5 frames at 5~Hz \\
        Agent / map capacity & 96 / 256 tokens \\
        Continuous action & Longitudinal jerk and tire steering-angle rate \\
        Normalized action support & $(0,1)^2$ \\
        Physical action ranges & Jerk $[-8,5]$~$\mathrm{m/s^3}$; steering rate $[-0.8,0.8]$~$\mathrm{rad/s}$ \\
        Vehicle-state constraints & Acceleration cap $4.0$~$\mathrm{m/s^2}$; steering-angle cap $\pi/3$~rad \\
        Training / evaluation action & Beta sample / analytic Beta mode \\
        \bottomrule
    \end{tabularx}
\end{table}

\paragraph{Evaluation and Benchmarks.}We evaluate \teacher on three nuPlan community splits: Val14 (1,118 scenarios), Test14-hard (272), and Test14-random (261)~\citep{caesar2021nuplan,dauner2023pdmclosed,cheng2024plantf}. Each 15-s scenario is evaluated in closed-loop non-reactive (NR) and reactive (R) modes. Background actors follow their logged trajectories in NR, while eligible vehicles are controlled by IDM in R and the remaining actors stay on log replay. We report the standard nuPlan closed-loop score (CLS), which measures progress, safety, compliance, and comfort on a 0--100 scale. \teacher and \teacher-TTS share the same checkpoint. \teacher executes the mode of its Beta policy, whereas \teacher-TTS applies the value-guided test-time action search described in Section~\ref{sec:tts-method}. The two settings therefore differ only in inference-time action selection and require no retraining.

\subsubsection{Closed-Loop Comparison}
\label{sec:main-results-teacher}

Table~\ref{tab:public-comparison} compares \teacher with published methods.
\teacher obtains an unweighted mean score of 93.01 across the six evaluations and exceeds Log-Replay in every non-reactive and reactive setting, showing that closed-loop RL can learn driving behavior beyond the demonstrations that seed its training worlds. It also exceeds CaRL and GigaFlow, the two prior methods trained without human data, on every setting they report.
Applying value-guided test-time action search to the same checkpoint raises the mean score by 0.56 points to 93.57, improving five of the six settings while remaining essentially unchanged on Test14-hard R (89.15 vs. 89.18). Detailed component metrics for \teacher are reported in Appendix~\ref{app:detailed-nuplan}, Table~\ref{tab:appendix-single-ego-results}.

\begin{table*}[t]
  \caption{\textbf{Closed-Loop Performance on the nuPlan~\cite{caesar2021nuplan} Benchmark.} 
  ``Human'' indicates whether human demonstrations are used as training supervision. 
  \teacher-TTS applies value-guided test-time action search to the same checkpoint as \teacher. Best scores are \textbf{bolded}; second-best scores are \underline{underlined}. 
  } 
  \label{tab:public-comparison}
  \centering
  \footnotesize
  \renewcommand{\arraystretch}{1.2}
  \setlength{\tabcolsep}{3pt}
  \begin{tabularx}{0.96\textwidth}{@{}l|c|*{6}{>{\centering\arraybackslash}X}@{}}
    \toprule
    \multirow{2}{*}[-0.15ex]{\textbf{Method}} & \multirow{2}{*}[-0.15ex]{\textbf{Human}} & \multicolumn{2}{c}{\textbf{Val14} $\uparrow$} & \multicolumn{2}{c}{\textbf{Test14-hard} $\uparrow$} & \multicolumn{2}{c}{\textbf{Test14-random} $\uparrow$} \\
    \cmidrule(lr){3-4} \cmidrule(lr){5-6} \cmidrule(lr){7-8}
    & & \textbf{NR} & \textbf{R} & \textbf{NR} & \textbf{R} & \textbf{NR} & \textbf{R} \\
    \midrule
    Log-Replay & \xmark & 93.53 & 80.32 & 85.96 & 68.80 & 94.03 & 75.86 \\
    \midrule
    IDM~\citep{treiber2000idm} & \xmark & 75.60 & 77.33 & 56.15 & 62.26 & 70.39 & 74.42 \\
    PDM-Closed~\citep{dauner2023pdmclosed} & \cmark & 92.84 & 92.12 & 65.08 & 75.19 & 90.05 & 91.63 \\
    \midrule
    UrbanDriver~\citep{scheel2021urbandriver} & \cmark & 68.57 & 64.11 & 50.40 & 49.95 & 51.83 & 67.15 \\
    PlanTF~\citep{cheng2024plantf} & \cmark & 84.27 & 76.95 & 69.70 & 61.61 & 85.62 & 79.58 \\
    PLUTO~\citep{cheng2024pluto} & \cmark & 89.04 & 80.01 & -- & -- & -- & -- \\
    Diffusion Planner~\citep{zheng2025diffusionplanner} & \cmark & 89.87 & 82.80 & 75.99 & 69.22 & 89.19 & 82.93 \\
    Flow Planner~\citep{tan2025flowplanner} & \cmark & 90.43 & 83.31 & 76.47 & 70.42 & 89.88 & 82.93 \\
    Pi-DiMT~\citep{zhou2026pidimt} & \cmark & 89.66 & 82.95 & 77.08 & 71.02 & 92.45 & 87.09 \\
    Plan-R1~\citep{tang2026planr1} & \cmark & 88.98 & 87.69 & 77.45 & 77.20 & 91.23 & 90.04 \\
    PlannerRFT~\citep{li2026plannerrft} & \cmark & 89.96 & 84.46 & 77.16 & 72.21 & 90.76 & 85.80 \\
    \midrule
    CaRL~\citep{jaeger2025carl} & \xmark & 93.87 & 93.12 & -- & 85
    & -- & -- \\
    GigaFlow~\citep{cusumano2025selfplay} & \xmark & -- & 93.80 & -- & -- & -- & -- \\
    \midrule
    \rowcolor[HTML]{FFE0CC}
    \multicolumn{8}{c}{\textbf{\textit{Our Methods}}} \\
    \midrule
    \textbf{\teacher} & \xmark & \underline{95.16} & \underline{94.25} & \underline{89.97} & \textbf{89.18} & \underline{94.50} & \underline{95.00} \\
    \textbf{\teacher-TTS} & \xmark & \textbf{95.54} & \textbf{94.53} & \textbf{91.13} & \underline{89.15} & \textbf{95.93} & \textbf{95.12} \\
    \bottomrule
  \end{tabularx}
\end{table*}

\subsubsection{Test-Time Scaling}
\label{sec:tts-experiments}

We next isolate the effect of additional inference computation while keeping the \teacher checkpoint fixed. All runs use lookahead $L=5$ and switching margin of $\delta=0.03$. We vary the number of candidates $N$ from 8 to 64 and evaluate on the six settings of Section~\ref{sec:main-results-teacher}.

\begin{table*}[t]
  \caption{\textbf{Test-time Scaling of a Fixed \teacher Checkpoint}. \teacher directly executes the Beta mode, whereas \teacher-TTS evaluates policy-supported action candidates using a five-step rollout and a switching margin. Mean is the unweighted average over the six evaluations, and $\Delta$ is measured relative to \teacher.}
  \label{tab:tts-scaling}
  \centering
  \footnotesize
  \setlength{\tabcolsep}{1.3pt}
  \begin{tabular*}{\textwidth}{@{\extracolsep{\fill}}l|cccccc|cc@{}}
    \toprule
    \multirow{2}{*}[-0.15ex]{\textbf{Setting}}
    & \multicolumn{2}{c}{\textbf{Val14} $\uparrow$}
    & \multicolumn{2}{c}{\textbf{Test14-hard} $\uparrow$}
    & \multicolumn{2}{c|}{\textbf{Test14-random} $\uparrow$}
    & \multirow{2}{*}[-0.15ex]{\textbf{Mean} $\uparrow$}
    & \multirow{2}{*}[-0.15ex]{$\boldsymbol{\Delta}$ $\uparrow$} \\
    \cmidrule(lr){2-3} \cmidrule(lr){4-5} \cmidrule(lr){6-7}
    & \textbf{NR} & \textbf{R} & \textbf{NR} & \textbf{R} & \textbf{NR} & \textbf{R} & & \\
    \midrule
    \teacher & 95.16 & 94.25 & 89.97 & \textbf{89.18} & 94.50 & 95.00 & 93.01 & 0.00 \\
    \teacher-TTS ($N=8$) & 95.17 & 94.49 & 89.97 & 89.05 & 94.92 & 95.11 & 93.12 & +0.11 \\
    \teacher-TTS ($N=16$) & 95.22 & 94.38 & 89.98 & 88.97 & 95.29 & 94.98 & 93.13 & +0.13 \\
    \teacher-TTS ($N=32$) & 95.49 & 94.43 & 90.95 & 88.85 & 95.83 & 94.84 & 93.40 & +0.39 \\
    \teacher-TTS ($N=64$) & \textbf{95.54} & \textbf{94.53} & \textbf{91.13} & 89.15 & \textbf{95.93} & \textbf{95.12} & \textbf{93.57} & \textbf{+0.56} \\
    \bottomrule
  \end{tabular*}
\end{table*}

As shown in Table~\ref{tab:tts-scaling}, the mean score generally increases with the candidate budget, reaching 93.57 with 64 candidates. The largest gains occur on the non-reactive Test14 splits: Test14-hard NR improves by 1.16 points, while Test14-random NR improves by 1.43 points. Overall, the scaling trend is consistent with the intended role of test-time search as a local policy-improvement operator. A larger candidate set increases the chance of finding a first action whose short rollout avoids an unfavorable local consequence while retaining a high-value continuation.

\subsection{\student}
\paragraph{Training Dataset.}
For training, we use the \cmtt{navtrain} split from NAVSIM~\citep{Dauner2024navsim}, which is built on nuPlan~\citep{karnchanachari2024nuplan} and contains 100K interactive real-world scenarios. Instead of using logged human driving trajectories, the student is supervised by trajectories accumulated from \teacher{} rollouts.
We additionally use 237K out-of-distribution (OOD) simulation scenes from SimScale~\citep{tian2026simscale} for data scaling, and denote the resulting variant as \student-Scale.

\paragraph{Training Details.}
\student{} takes as input the four camera views \cmtt{CAM\_F0}, \cmtt{CAM\_B0}, \cmtt{CAM\_L0}, and \cmtt{CAM\_R0}. The planner comprises 64 trajectory proposals, a 256-dimensional planning representation, and a 4-layer proposal decoder. Following DrivoR~\citep{kirby2026drivor}, camera features are compressed using 16 register tokens per camera. During training, the DriveVFM backbone remains frozen, with only rank-32 Q/V LoRA adapters~\citep{hu2021lora} left trainable. We train \student{} for 25 epochs with a batch size of 256, using AdamW~\citep{loshchilov2017adamw} and an initial learning rate of $2\times10^{-4}$ decayed to zero via cosine annealing. All main-table DriveZero variants use the DriveVFM ViT-L backbone; results with ViT-S and ViT-B are reported in Appendix \ref{app:student-backbone-scaling}.

\paragraph{Benchmarks.}
We evaluate the camera-only student on three complementary benchmarks. NAVSIMv1 \cmtt{navtest}~\citep{Dauner2024navsim}, built from nuPlan as a subset of OpenScene, and NAVSIMv2 \cmtt{navhard}~\citep{Cao2025navsimv2} provide pseudo closed-loop evaluation under standard sensor-input protocols. On \cmtt{navtest}, we report the Predictive Driver Model Score (PDMS) and its no-collision, drivable-area compliance, time-to-collision, comfort, and ego-progress components. On \cmtt{navhard}, we report the two-stage Extended Predictive Driver Model Score (EPDMS) and its safety, compliance, progress, and planning-quality components; its second stage evaluates robustness to Gaussian-Splatting-based ego-state perturbations. We additionally evaluate zero-shot transfer on HUGSIM~\citep{zhou2024hugsim}, a true closed-loop benchmark reconstructed from scenes in KITTI-360~\citep{kitti360}, nuScenes~\citep{nuscenes}, PandaSet~\citep{pandaset}, and Waymo~\citep{waymo}. We report route completion and HD-Score over Easy, Medium, Hard, and Extreme scenarios. Zero-shot means no HUGSIM-specific finetuning is used.

\subsubsection{Main Results of the End-to-End Student}
\label{sec:main-results-student}
\begin{table*}[t]
\centering

\caption{
\textbf{Pseudo Closed-Loop Performance on the NAVSIMv1
navtest~\cite{Dauner2024navsim} benchmark.}
Input: GT = ground-truth symbolic inputs; C = camera; L = LiDAR. ``Human'' indicates whether human demonstrations are used as training supervision.
``-Scale'' denotes scaling up the training set with simulation data
from SimScale~\cite{tian2026simscale}. Best scores are \textbf{bolded}; second-best scores are \underline{underlined}. 
}

\footnotesize
\renewcommand{\arraystretch}{1.2}
\setlength{\tabcolsep}{5pt}

\begin{tabular}{l|c|c|ccccc|c}
\toprule

\textbf{Method}
& \textbf{Input}
& \textbf{Human}
& \textbf{NC} $\uparrow$
& \textbf{DAC} $\uparrow$
& \textbf{TTC} $\uparrow$
& \textbf{C} $\uparrow$
& \textbf{EP} $\uparrow$
& \textbf{PDMS} $\uparrow$ \\

\midrule

\textcolor{gray}{PDM-Closed}
& \textcolor{gray}{GT}
& \textcolor{gray}{\cmark}
& \textcolor{gray}{94.6}
& \textcolor{gray}{99.8}
& \textcolor{gray}{89.9}
& \textcolor{gray}{86.9}
& \textcolor{gray}{99.9}
& \textcolor{gray}{89.1} \\

\textcolor{gray}{Human Driver}
& \textcolor{gray}{GT}
& \textcolor{gray}{\cmark}
& \textcolor{gray}{100}
& \textcolor{gray}{100}
& \textcolor{gray}{100}
& \textcolor{gray}{99.9}
& \textcolor{gray}{87.5}
& \textcolor{gray}{94.8} \\
\midrule

TransFuser~\cite{kashyap2022transfuser}
& C \& L
& \cmark
& 97.7 & 92.8 & 92.8 & \textbf{100} & 79.2 & 84.0 \\

DRAMA~\cite{yuan2024drama}
& C \& L
& \cmark
& 98.0 & 93.1 & 94.8 & \textbf{100} & 80.1 & 85.5 \\

VAD-v2~\cite{chen2024vadv2}
& C \& L
& \cmark
& 98.1 & 94.8 & 94.3 & \textbf{100} & 80.6 & 86.2 \\

DiffusionDrive~\cite{liao2025diffusiondrive}
& C \& L
& \cmark
& 98.2 & 96.2 & 94.7 & \textbf{100} & 82.2 & 88.1 \\

Hydra-MDP++~\cite{li2025hydramdp++}
& C \& L
& \cmark
& 98.6 & 98.6 & 95.1 & \textbf{100} & 85.7 & 91.0 \\

Centaur~\cite{sima2025centaur}
& C \& L
& \cmark
& \textbf{99.5} & 98.9 & \textbf{98.0} & \textbf{100} & 85.9 & 92.6 \\

DriveSuprim~\cite{yao2026drivesuprim}
& C \& L
& \cmark
& 98.6 & 98.6 & 95.5 & \textbf{100} & 91.3 & 93.5 \\

\midrule

UniAD~\cite{hu2023uniad}
& C
& \cmark
& 97.8 & 91.9 & 92.9 & \textbf{100} & 78.8 & 83.4 \\

PARA-Drive~\cite{weng2024paradrive}
& C
& \cmark
& 97.9 & 92.4 & 93.0 & 99.8 & 79.3 & 84.0 \\

Epona~\cite{zhang2025epona}
& C
& \cmark
& 97.9 & 95.1 & 93.8 & \underline{99.9} & 80.4 & 86.2 \\

OneVL~\cite{lu2026onevl}
& C
& \cmark
& 98.5 & 96.9 & 95.8 & \textbf{100} & 82.2 & 88.4 \\

DriveLaW~\cite{xia2026drivelaw}
& C
& \cmark
& 99.0 & 97.1 & 96.7 & \textbf{100} & 81.3 & 89.1 \\

AutoVLA~\cite{zhou2025autovla}
& C
& \cmark
& 98.4 & 95.6 & \textbf{98.0} & \underline{99.9} & 81.9 & 89.1 \\

DriveVLA-W0~\cite{li2025drivevla}
& C
& \cmark
& 98.7 & 99.1 & 95.3 & 99.3 & 83.3 & 90.2 \\

Qwen-Drive-1.0~\cite{zhou2026qwendrive}
& C
& \cmark
& 98.6 & 98.2 & 95.9 & \textbf{100} & 84.8 & 90.7 \\

ReCogDrive~\cite{li2025recogdrive}
& C
& \cmark
& 97.9 & 97.3 & 94.9 & \textbf{100} & 87.3 & 90.8 \\

R2SE~\cite{liu2025reinforced}
& C
& \cmark
& 99.0 & 97.9 & 96.4 & \textbf{100} & 86.8 & 91.6 \\

iPad~\cite{guo2025ipad}
& C
& \cmark
& 98.6 & 98.3 & 94.9 & \textbf{100} & 88.0 & 91.7 \\

DriveFine~\cite{dang2026drivefine}
& C
& \cmark
& 98.8 & 98.6 & 96.2 & \textbf{100} & 86.9 & 91.8 \\

DrivoR-Scale~\cite{kirby2026drivor}
& C
& \cmark
& 99.1 & \underline{99.2} & 96.9 & \textbf{100} & 91.6 & 94.6 \\
\midrule

\rowcolor[HTML]{FFE0CC}
\multicolumn{9}{c}{\textbf{\textit{Our Methods}}} \\
\midrule

\textcolor{gray}{\textbf{\teacher}}
& \textcolor{gray}{GT}
& \textcolor{gray}{\xmark}
& \textcolor{gray}{99.8}
& \textcolor{gray}{99.9}
& \textcolor{gray}{99.1}
& \textcolor{gray}{99.0}
& \textcolor{gray}{91.5}
& \textcolor{gray}{95.8} \\

\textbf{\student}
&C
& \xmark
& 99.0 & \underline{99.2} & 96.0 & \textbf{100} & \textbf{93.1} & \underline{94.8} \\

\textbf{\student-Scale}
&C
& \xmark
& \underline{99.2} & \textbf{99.4} & \underline{97.3} & \underline{99.9} & \underline{92.5} & \textbf{95.3} \\

\bottomrule


\end{tabular}

\label{tab:navtest}

\end{table*}

\begin{table*}[t]
\centering
\caption{
\textbf{Pseudo Closed-Loop Performance on the NAVSIMv2 navhard~\cite{Cao2025navsimv2} Benchmark. }
$^\ast$ marks methods using ground-truth symbolic inputs; all others use camera inputs. ``Human'' indicates whether human demonstrations are used as training supervision, and ``S.'' indicates per-stage EPDM score.
``-Scale'' denotes scaling up the training set with simulation data
from SimScale~\cite{tian2026simscale}. Best scores are \textbf{bolded}; second-best scores are \underline{underlined}.
}
\label{tab:navhard}

\renewcommand{\arraystretch}{1.1}
\setlength{\tabcolsep}{5pt}

\resizebox{\textwidth}{!}{%
\begin{tabular}{l|c|c|cccc|ccccc|c|c}
\toprule

\textbf{Method}
& \textbf{Human}
& \textbf{Stage}
& \textbf{NC} $\uparrow$
& \textbf{DAC} $\uparrow$
& \textbf{DDC} $\uparrow$
& \textbf{TLC} $\uparrow$
& \textbf{EP} $\uparrow$
& \textbf{TTC} $\uparrow$
& \textbf{LK} $\uparrow$
& \textbf{HC} $\uparrow$
& \textbf{EC} $\uparrow$
& \textbf{S.} $\uparrow$
& \textbf{EPDMS} $\uparrow$ \\

\midrule

\textcolor{gray}{} & \textcolor{gray}{} & \textcolor{gray}{S1}
& \textcolor{gray}{94.4}
& \textcolor{gray}{98.8}
& \textcolor{gray}{100}
& \textcolor{gray}{99.5}
& \textcolor{gray}{100}
& \textcolor{gray}{93.5}
& \textcolor{gray}{99.3}
& \textcolor{gray}{87.7}
& \textcolor{gray}{36.0}
& \textcolor{gray}{-}
& \\

\multirow{-2}{*}{\textcolor{gray}{PDM-Closed{$^\ast$}~\cite{dauner2023pdmclosed}}
}
&
\multirow{-2}{*}{\textcolor{gray}{\cmark}}
&
\textcolor{gray}{S2}
&
\textcolor{gray}{90.5}
&
\textcolor{gray}{90.6}
&
\textcolor{gray}{95.4}
&
\textcolor{gray}{98.4}
&
\textcolor{gray}{100}
&
\textcolor{gray}{86.6}
&
\textcolor{gray}{74.2}
&
\textcolor{gray}{91.9}
&
\textcolor{gray}{29.7}
&
\textcolor{gray}{-}
&
\multirow{-2}{*}{\textcolor{gray}{56.6}}
\\

\midrule

& & S1
& 96.2 & 79.6 & 99.1 & 99.6
& 84.1 & 95.1 & 94.2 & 97.6 & 79.1
& - & \\

\multirow{-2}{*}{LTF~\cite{kashyap2022transfuser}}
& \multirow{-2}{*}{\cmark}
& S2
& 77.8 & 70.2 & 84.3 & 98.1
& 85.1 & 75.7 & 45.4 & 95.7 & 76.0
& -
& \multirow{-2}{*}{25.1} \\

\cmidrule{3-14}

& & S1
& 96.2 & 92.4 & 95.7 & 99.6
& 83.8 & 96.0 & 94.7 & 96.4 & 60.9
& - & \\

\multirow{-2}{*}{NavFormer~\cite{li2026worldengine}}
& \multirow{-2}{*}{\cmark}
& S2
& 85.7 & 81.0 & 83.5 & 97.6
& 90.1 & 82.4 & 48.2 & 94.9 & 48.4
& -
& \multirow{-2}{*}{34.1} \\

\cmidrule{3-14}

& & S1
& 97.1 & 94.4 & 98.8 & 99.8
& 83.9 & 96.9 & 94.7 & 96.4 & 66.2
& - & \\

\multirow{-2}{*}{RAP~\cite{feng2025rap}}
& \multirow{-2}{*}{\cmark}
& S2
& 83.2 & 83.9 & 87.4 & 98.0
& 86.9 & 80.4 & 52.3 & 95.2 & 52.4
& -
& \multirow{-2}{*}{39.6} \\

\cmidrule{3-14}

& & S1
& 99.6 & 98.0 & 99.4 & 99.3
& 79.7 & 99.3 & 94.9 & 97.1 & 58.2
& - & \\

\multirow{-2}{*}{GuideFlow~\cite{liu2026guideflow}}
& \multirow{-2}{*}{\cmark}
& S2
& 91.4 & 89.5 & 95.2 & 98.9
& 77.5 & 89.6 & 52.6 & 93.6 & 51.0
& -
& \multirow{-2}{*}{51.5} \\

\cmidrule{3-14}

& & S1
& 99.6 & 99.1 & 99.9 & 100.0
& 69.6 & 99.6 & 95.8 & 95.6 & 28.4
& - & \\

\multirow{-2}{*}{SimScale~\cite{tian2026simscale}}
& \multirow{-2}{*}{\cmark}
& S2
& 94.5 & 94.2 & 95.8 & 99.2
& 75.8 & 92.8 & 60.1 & 96.1 & 43.2
& -
& \multirow{-2}{*}{53.2} \\

\cmidrule{3-14}

& & S1
& 99.1 & 98.2 & 99.3 & 99.8
& 75.4 & 98.7 & 94.9 & 97.6 & 70.2
& - & \\

\multirow{-2}{*}{DrivoR-Scale~\cite{kirby2026drivor}}
& \multirow{-2}{*}{\cmark}
& S2
& 92.3 & 91.6 & 97.3 & 99.1
& 75.7 & 90.6 & 56.1 & 98.4 & 44.7
& -
& \multirow{-2}{*}{\underline{54.6}} \\

\midrule

& & S1
& 98.9 & 97.6 & 100.0 & 100.0
& 66.7 & 98.9 & 96.2 & 96.7 & 44.0
& - & \\

\multirow{-2}{*}{ZTRS~\cite{li2025ztrs}}
& \multirow{-2}{*}{\xmark}
& S2
& 91.1 & 90.4 & 95.8 & 99.0
& 63.6 & 89.8 & 60.4 & 97.6 & 66.1
& -
& \multirow{-2}{*}{48.1} \\

\cmidrule{3-14}

& & S1
& 99.4 & 95.8 & 99.4 & 99.8
& 68.1 & 99.6 & 91.8 & 97.6 & 49.8
& 77.8 & \\

\multirow{-2}{*}{GigaPixel~\cite{rowe2026gigapixel}}
& \multirow{-2}{*}{\xmark}
& S2
& 93.7 & 92.9 & 96.0 & 98.9
& 62.6 & 90.7 & 60.2 & 98.2 & 58.8
& \underline{63.5}
& \multirow{-2}{*}{50.1} \\

\midrule

\rowcolor[HTML]{FFE0CC}
\multicolumn{14}{c}{\textbf{\textit{Our Methods}}} \\
\midrule

\textcolor{gray}{} & \textcolor{gray}{} & \textcolor{gray}{S1}
& \textcolor{gray}{99.1}
& \textcolor{gray}{99.6}
& \textcolor{gray}{100.0}
& \textcolor{gray}{99.6}
& \textcolor{gray}{91.1}
& \textcolor{gray}{98.7}
& \textcolor{gray}{99.3}
& \textcolor{gray}{90.9}
& \textcolor{gray}{18.2}
& \textcolor{gray}{84.1}
& \\

\multirow{-2}{*}{\textcolor{gray}{\textbf{\teacher}$^\ast$}}
& \multirow{-2}{*}{\textcolor{gray}{\xmark}}
& \textcolor{gray}{S2}
& \textcolor{gray}{94.4}
& \textcolor{gray}{92.0}
& \textcolor{gray}{96.4}
& \textcolor{gray}{98.1}
& \textcolor{gray}{89.1}
& \textcolor{gray}{89.1}
& \textcolor{gray}{73.8}
& \textcolor{gray}{88.5}
& \textcolor{gray}{10.9}
& \textcolor{gray}{65.4}
& \multirow{-2}{*}{\textcolor{gray}{55.6}} \\

\cmidrule{3-14}

& & S1
& 98.7 & 98.9 & 99.9 & 99.8
& 79.5 & 98.7 & 95.8 & 97.6 & 51.6
&\textbf{84.1} & \\

\multirow{-2}{*}{\textbf{\student}}
& \multirow{-2}{*}{\xmark}
& S2
& 90.5 & 89.2 & 94.6 & 98.6
& 81.2 & 88.3 & 57.2 & 94.5 & 41.3
& 60.8 & \multirow{-2}{*}{51.5} \\

\cmidrule{3-14}

& & S1
& 98.7 & 99.3 & 100.0 & 99.3
& 76.5 & 98.4 & 95.3 & 97.1 & 44.4
& \underline{82.3} & \\

\multirow{-2}{*}{\textbf{\student-Scale}}
& \multirow{-2}{*}{\xmark}
& S2
& 94.6 & 95.6 & 97.7 & 98.4
& 76.2 & 92.7 & 59.9 & 96.1 & 44.0
& \textbf{69.1} & \multirow{-2}{*}{\textbf{57.1}}\\

\bottomrule  


\end{tabular}%
}

\end{table*}

\begin{table}[t]
\centering
\caption{
\textbf{True Closed-Loop Performance on the HUGSIM~\cite{zhou2024hugsim} Benchmark.}
Zero-shot generalization using the \student model from the
NAVSIMv1 evaluation. ``Human'' indicates whether human demonstrations are used as training supervision. ``-Scale'' denotes scaling up the training set with simulation data
from SimScale~\cite{tian2026simscale}.
Best scores are \textbf{bolded}; second-best scores are \underline{underlined}.
}
\footnotesize
\renewcommand{\arraystretch}{1.2}
\setlength{\tabcolsep}{5.5pt}
\begin{tabular}{l|c|cccc|c|cccc|c}
\toprule
\multirow{2}{*}[-0.15ex]{\textbf{Method}} &
\multirow{2}{*}[-0.15ex]{\textbf{Human}} &
\multicolumn{5}{c|}{\textbf{RC} $\uparrow$} &
\multicolumn{5}{c}{\textbf{HD-Score} $\uparrow$} \\
\cmidrule(lr){3-7} \cmidrule(lr){8-12}
& &
\textbf{E} & \textbf{M} & \textbf{H} & \textbf{X} & \textbf{Avg.} &
\textbf{E} & \textbf{M} & \textbf{H} & \textbf{X} & \textbf{Avg.} \\
\midrule
LTF~\cite{kashyap2022transfuser}
& \cmark
& 67.8 & 35.1 & 26.2 & 40.5 & 38.9
& 58.9 & 18.0 & 9.8 & 25.9 & 23.7 \\
VAD~\cite{jiang2023vad}
& \cmark
& 51.3 & 31.1 & 25.3 & 26.5 & 31.4
& 36.3 & 9.5 & 8.0 & 11.5 & 13.4 \\
UniAD~\cite{hu2023uniad}
& \cmark
& 78.4 & 60.5 & \underline{33.6} & 17.8 & 45.9
& 64.9 & 45.8 & \underline{20.6} & 6.6 & 32.7 \\
DrivoR-Scale~\cite{kirby2026drivor}
& \cmark
& 70.4 & 54.4 & 28.2 & 39.2 & 46.4
& 64.8 & 50.3 & 16.8 & 25.8 & 38.1 \\
\midrule
GigaPixel~\cite{rowe2026gigapixel}
& \xmark
& 78.6 & 60.7 & 32.7 & 35.4 & 50.1
& 67.4 & \underline{51.9} & 19.1 & 21.6 & 38.5 \\
\midrule
\rowcolor[HTML]{FFE0CC}
\multicolumn{12}{c}{\textbf{\textit{Our Methods}}} \\
\midrule
\textbf{\student}
& \xmark
& \underline{85.5} & \underline{63.7} & 31.7 & \textbf{43.3} & \underline{53.9}
& \underline{78.5} & 47.2 & 16.0 & \textbf{28.7} & \underline{39.4} \\
\textbf{\student-Scale}
& \xmark
& \textbf{87.5} & \textbf{64.9} & \textbf{34.6} & \underline{39.8} & \textbf{54.5}
& \textbf{81.9} & \textbf{60.5} & \textbf{24.7} & \underline{27.6} & \textbf{46.6} \\
\bottomrule
\end{tabular}
\label{tab:hugsim}
\end{table}

\paragraph{Pseudo Closed-Loop Leaderboard: Navtest.}
As shown in Table~\ref{tab:navtest}, the RL-trained \teacher{} achieves 95.8 PDMS with 4-s rollouts from ground-truth symbolic inputs, outperforming Human Driver (94.8) and PDM-Closed (89.1). It combines near-human safety (99.8 NC and 99.9 DAC) with substantially higher driving efficiency, improving EP from 87.5 to 91.5. High-quality, goal-controllable supervision from \teacher{}, together with DriveVFM, enables the camera-only \student{} to reach 94.8 PDMS without human trajectory supervision, outperforming all prior camera-only and camera--LiDAR fusion methods. Scaling with OOD simulation data from SimScale~\cite{tian2026simscale} further improves \student-Scale to 95.3 PDMS, providing direct evidence of the scalability of our framework. \student-Scale also preserves strong safety and driving efficiency, with NC and DAC above 99 and EP at 92.5. To the best of our knowledge, \textbf{this is the first end-to-end model to outperform the Human Driver on the NAVSIMv1 navtest leaderboard.}

\paragraph{Pseudo Closed-Loop Leaderboard: Navhard.}
As shown in Table~\ref{tab:navhard}, \student{} achieves an EPDMS of 51.5, outperforming all methods trained exclusively on the \cmtt{navtrain} split. It also exceeds GigaPixel~\cite{rowe2026gigapixel}, another RL-Teacher-based approach, by 1.4 points, even though the student is trained open-loop on logged frames and never enters a simulator.
Scaling with SimScale~\cite{tian2026simscale} data raises the combined score by 5.6 points to 57.1: the Stage 1 score decreases by 1.8 points, while the more challenging Stage 2 score improves by 8.3 points. This establishes a new state of the art, surpassing both DrivoR-Scale~\cite{kirby2026drivor}, which also uses SimScale data, and PDM-Closed~\cite{dauner2023pdmclosed}, despite the latter using ground-truth symbolic inputs.

\paragraph{True Closed-Loop Leaderboard: HUGSIM.}
As shown in Table~\ref{tab:hugsim}, by data scaling, the \student-Scale{} delivers the strongest zero-shot closed-loop performance. It achieves state-of-the-art HD-Score on the Easy, Medium, and Hard tiers, resulting in an average HD-Score of 46.6. This improves on the previous best result from GigaPixel~\cite{rowe2026gigapixel} by 8.1 points (46.6 vs.~38.5). We attribute this gain to the more plausible and diverse trajectories generated by \teacher on the OOD
 SimScale~\cite{tian2026simscale} data, which enable the Student to exploit the additional data more effectively.

\subsubsection{Ablation Studies}
\label{sec:student-ablations}

Table~\ref{tab:student-ablations} summarizes the three controlled studies. All ablations use a ViT-S backbone and the \cmtt{navtrain} split without SimScale data, and are evaluated on NAVSIMv1 \cmtt{navtest}. We report the PDMS and five of its components so that changes in the aggregate score can be traced to specific aspects of driving quality.

\begin{table}[H]
  \caption{\textbf{DriveVFM and \student{} ablations on the NAVSIMv1 navtest~\cite{Dauner2024navsim} benchmark.}}
  \label{tab:student-ablations}
  \centering
  \footnotesize
  \setlength{\tabcolsep}{3.5pt}
  \begin{tabular*}{\textwidth}{@{\extracolsep{\fill}}ll|ccccc|c}
    \toprule
    \textbf{Study} & \textbf{Configuration} & \textbf{NC} $\uparrow$ & \textbf{DAC} $\uparrow$ & \textbf{EP} $\uparrow$ & \textbf{TTC} $\uparrow$ & \textbf{Comfort} $\uparrow$ & \textbf{PDMS} $\uparrow$ \\
    \midrule
    \multirow{5}{*}{Visual Backbone}
      & DA3 Small~\citep{lin2025depthanything3} & 99.10 & 98.84 & 89.68 & 96.21 & 99.93 & 93.31 \\
      & DA2 ViT-S~\citep{yang2024depthanythingv2} & 98.57 & \textbf{99.21} & 91.44 & 95.13 & 99.96 & 93.72 \\
      & DINOv2 ViT-S~\citep{oquab2023dinov2} & 98.86 & 99.07 & 91.63 & 95.46 & \textbf{99.98} & 93.88 \\
      & DINOv3 ViT-S~\citep{simeoni2025dinov3} & 98.93 & 99.01 & 91.31 & 95.77 & 99.97 & 93.88 \\
      & \textbf{DriveVFM ViT-S} & \textbf{99.17} & 99.14 & \textbf{91.79} & \textbf{96.29} & \textbf{99.98} & \textbf{94.41} \\
    \midrule
    \multirow{3}{*}{DriveVFM Teachers}
      & DINOv3 + SigLIP2~\citep{simeoni2025dinov3,tschannen2025siglip2} & 98.67 & 98.98 & 91.55 & 95.27 & \textbf{99.98} & 93.69 \\
      & + SAM~\citep{kirillov2023sam} & 98.79 & 99.11 & 91.44 & 96.06 & 99.97 & 94.10 \\
      & +DA2~\citep{yang2024depthanythingv2} = Full & \textbf{99.17} & \textbf{99.14} & \textbf{91.79} & \textbf{96.29} & \textbf{99.98} & \textbf{94.41} \\
    \midrule
    \multirow{3}{*}{\student{} Supervision}
      & Human trajectories & 98.76 & 98.95 & \textbf{91.99} & 95.30 & \textbf{99.98} & 93.92 \\
      & \teacher{} trajectories & 98.60 & 98.81 & 91.64 & 95.27 & 99.94 & 93.61 \\
      & \textbf{\teacher{} + goal aug.} & \textbf{99.17} & \textbf{99.14} & 91.79 & \textbf{96.29} & \textbf{99.98} & \textbf{94.41} \\
    \bottomrule
  \end{tabular*}
\end{table}

\paragraph{Visual foundation model.}
The first block of Table~\ref{tab:student-ablations} investigates whether DriveVFM provides a stronger visual representation for end-to-end driving than existing general-purpose foundation models. We replace DriveVFM with DINOv2 and DINOv3 backbones of comparable capacity while keeping the trajectory head, camera inputs, training data, optimization schedule, and evaluation protocol unchanged. This controlled comparison isolates the contribution of visual pretraining from that of the downstream planning architecture. We report both the overall planning score and its safety, progress, and comfort components to determine whether any advantage of DriveVFM extends beyond a single aggregate metric. DriveVFM achieves a PDMS of 94.41, outperforming the DINOv3 ViT-S baseline at 93.88 by 0.53 points. This margin is notable on an already highly saturated benchmark.

\paragraph{Teacher contribution in DriveVFM.}
The second block of Table~\ref{tab:student-ablations} examines whether adding foundation-model teachers with complementary capabilities progressively improves DriveVFM. We begin with DINOv3 and SigLIP2 as the base teacher configuration, then add SAM to introduce object-aware and boundary-sensitive supervision, and finally add Depth Anything V2 to incorporate geometric cues. Each resulting backbone is evaluated through the same downstream \student{} trajectory-prediction task with all other training and evaluation settings unchanged. Adding SAM improves PDMS from 93.69 to 94.10 (+0.41), while further incorporating Depth Anything V2 raises it to 94.41 (+0.31). Together, the two additional teachers yield a 0.72-point improvement over the DINOv3 + SigLIP2 baseline, demonstrating cumulative gains from complementary supervision.

\paragraph{Trajectory distillation and goal augmentation.}
The final block of Table~\ref{tab:student-ablations} studies the central supervision components of \student{} through three progressively stronger settings: human trajectories, trajectories distilled from \teacher{}, and \teacher{} trajectories combined with goal augmentation. The first comparison replaces logged human futures with trajectories generated by \teacher{}. We then enable goal augmentation on top of \teacher{} distillation to measure its additional contribution. Although using \teacher{} trajectories alone yields a slightly lower PDMS than human supervision (93.61 versus 93.92), adding goal augmentation raises the score substantially to 94.41, outperforming \teacher-only supervision by 0.80 points and the human-trajectory baseline by 0.49 points. Goal augmentation is what makes teacher supervision surpass human supervision. It cannot be applied to the human-trajectory baseline, since a changed navigation goal would require a counterfactual human trajectory that logged data do not contain. The ability to generate goal-consistent supervision for arbitrary commands is therefore specific to the teacher.

\subsubsection{DriveVFM Feature Representation Visualization}
\label{app:student-feature-visualization}

\begin{figure}[H]
  \centering
  \includegraphics[width=\textwidth]{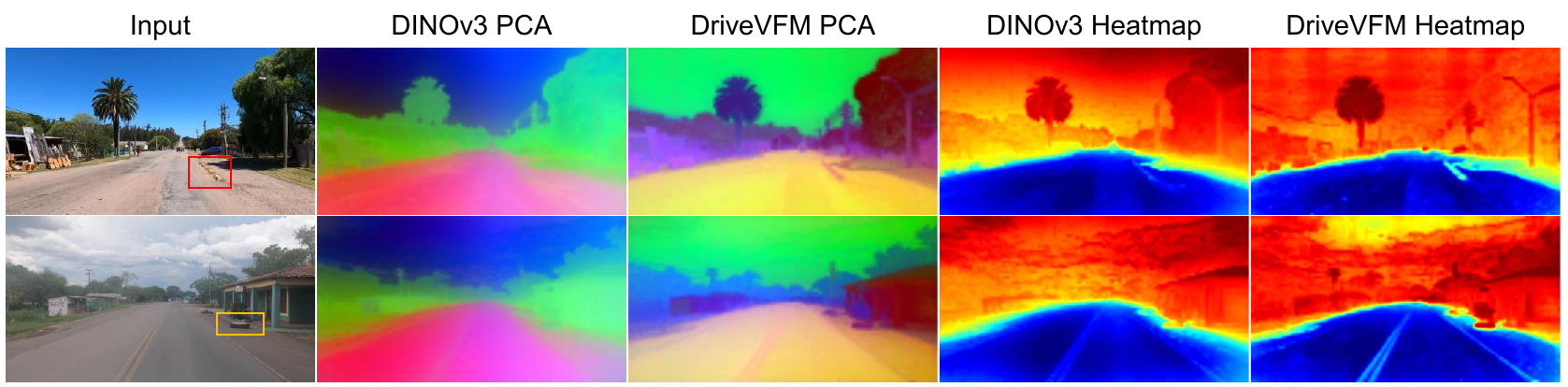}
  \caption{\textbf{Feature Representation Visualization.} From left to right: input images, PCA visualizations of DINOv3 and DriveVFM patch features, and their cosine-similarity heatmaps relative to the ground prototype. Blue indicates higher similarity to the ground and red indicates lower similarity. Boxes mark the long-tail obstacles of interest.}
  \label{fig:student-feature-visualization}
\end{figure}

Driving scenes contain long-tail obstacles that may not belong to common semantic categories but can still affect downstream planning. The visual backbone must therefore preserve the distinction between the road surface and such scene elements. We examine this property by comparing the dense patch representations of DINOv3 and DriveVFM in Figure~\ref{fig:student-feature-visualization}.

The second and third columns show PCA visualizations of the two representations. For each model, we use an off-the-shelf Mask2Former segmentation model~\citep{cheng2022mask2former} to identify the ground region and average the corresponding patch features to form a ground prototype. We then compute the cosine similarity between this prototype and every patch feature. The resulting heatmaps appear in the fourth and fifth columns, where blue denotes higher similarity to the ground and red denotes lower similarity.

Across both scenes, DriveVFM produces a more consistent response over the road surface and more clearly distinguishes long-tail obstacles and small objects from the ground than DINOv3.

\subsubsection{\student Planning Results Visualization}
\label{app:student-feature-visualization}

\begin{figure}[H]
  \centering
  \includegraphics[width=1\textwidth]{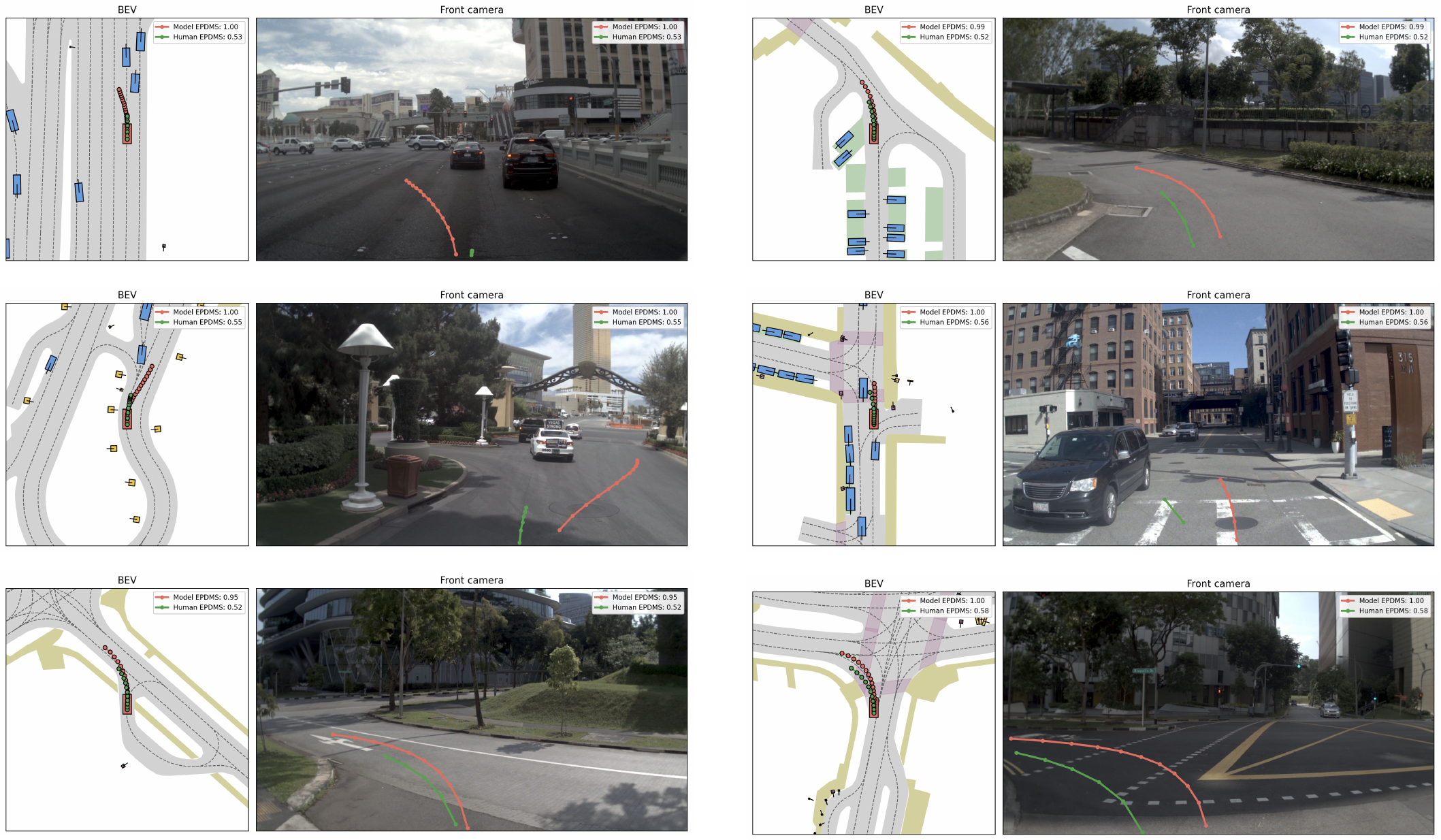}
  \caption{\textbf{Visualization of \textcolor{modelclr}{\student-Scale} vs. 
  \textcolor{humanclr}{Human Driver} in NAVSIM navtest~\cite{Dauner2024navsim} Benchmark.}}
  \label{fig:student-human-visualization}
\end{figure}

\paragraph{\student-Scale vs. Human Driver.} Figure~\ref{fig:student-human-visualization} compares the trajectories of DriveZero-Scale and the human driver on \cmtt{navtest} scenes. In these cases, the human driver hesitates or stays stopped although the road ahead is clear. DriveZero-Scale proceeds through the same scenes with a longer, smoother trajectory while keeping a safe distance from surrounding agents.

\begin{figure}[H]
  \centering
  \includegraphics[width=1\textwidth]{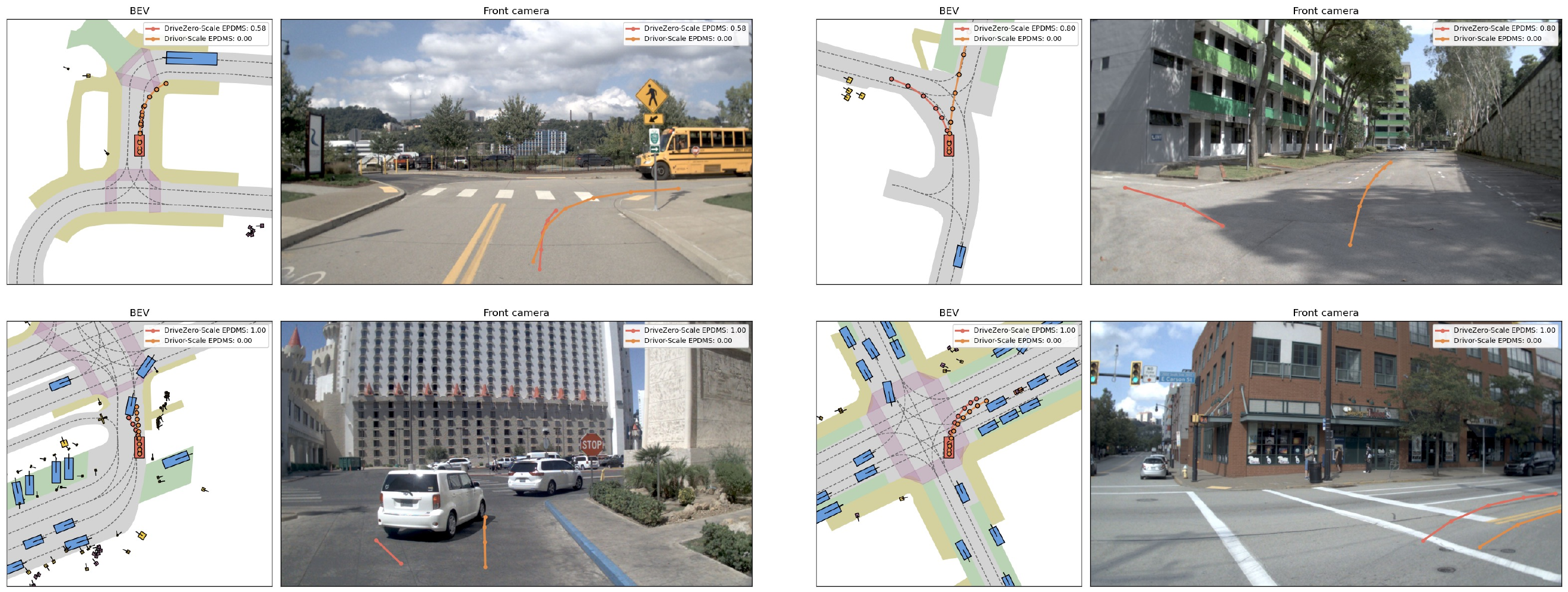}
  \caption{\textbf{Visualization of \textcolor{modelclr}{\student-Scale} vs. \textcolor{drivorclr}{Previous SOTA Model} in NAVSIM navtest~\cite{Dauner2024navsim} Benchmark.}}
  \label{fig:student-sota-navsim-visualization}
\end{figure}

\begin{figure}[H]
  \centering
  \begin{subfigure}[b]{\textwidth}
    \centering
    \includegraphics[width=\textwidth,trim=0 13 0 0,clip]{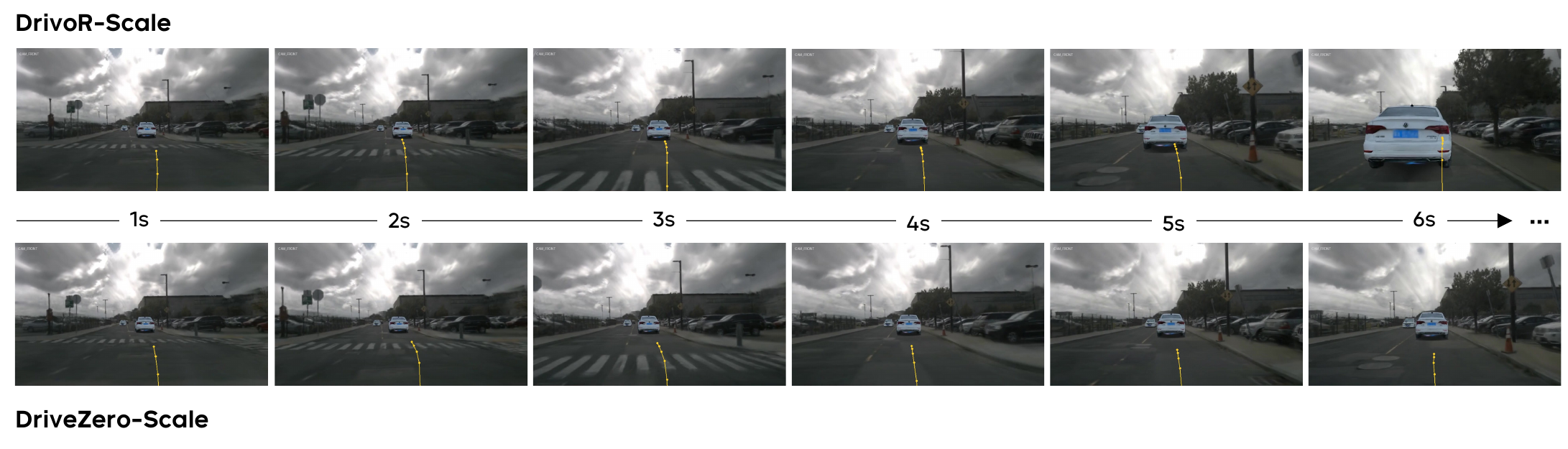}
    \label{fig:student-sota-hugsim-0}
  \end{subfigure}%
  \vspace{-7mm}
  {\color{gray!60}\rule{0.98\textwidth}{0.5pt}}
  \vspace{1mm}
  \begin{subfigure}[b]{\textwidth}
    \centering
    \includegraphics[width=\textwidth,trim=0 13 0 0,clip]{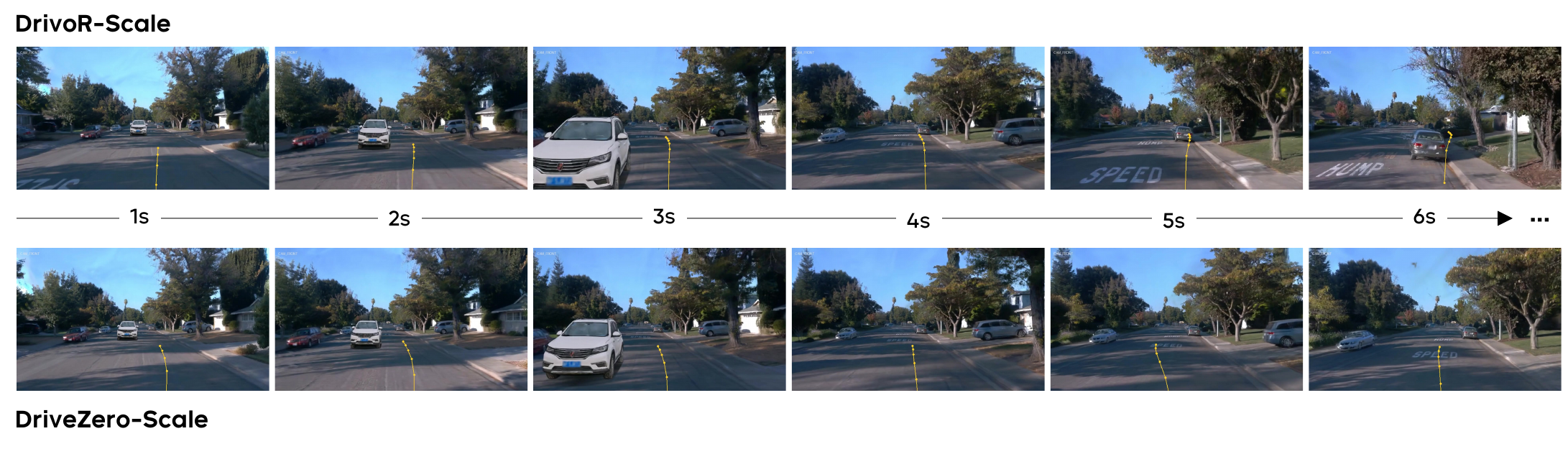}
    \label{fig:student-sota-hugsim-1}
  \end{subfigure}
  \caption{\textbf{Visualization of \textcolor{modelclr}{\student-Scale} vs. \textcolor{drivorclr}{Previous SOTA Model} in the HUGSIM~\cite{zhou2024hugsim} Benchmark.}}
  \label{fig:student-sota-hugsim-visualization}
\end{figure}

\paragraph{\student-Scale vs. Previous SOTA Model.} Figure~\ref{fig:student-sota-navsim-visualization} compares DriveZero-Scale with DrivoR-Scale, the strongest camera-only baseline in Table \ref{tab:navtest}, on \cmtt{navtest} scenes. In these scenes, the trajectory of DrivoR-Scale collides with a surrounding agent or leaves the drivable area, whereas DriveZero-Scale stays collision-free and within the road boundary. Figure~\ref{fig:student-sota-hugsim-visualization} shows two closed-loop rollouts from HUGSIM. In the first scene, the lead vehicle brakes suddenly. DrivoR-Scale keeps its speed and collides with it, whereas DriveZero-Scale decelerates in time and maintains a safe gap. In the second scene, an oncoming vehicle passes close to the ego lane. DrivoR-Scale swerves away from it and drifts onto the curb, while DriveZero-Scale stays in its lane with only a slight lateral adjustment. More closed-loop rollouts are available on our project page.

\subsection{Real-World Application}
\label{sec:realworld}

We have deployed \teacher{} on a small fleet of vehicles and validated it in real traffic.
For deployment, \teacher{} is trained on our own driving logs with self-play enabled.
As on nuPlan, the logs only seed the scenes and goals, and no human demonstration is used for training.
The structured state is built from auto-labeled perception ground truth, with recorded onboard perception outputs mixed in as degraded inputs, and extensive domain randomization is applied during training.
On the vehicle, the object detection, online mapping and navigation modules provide the surrounding agents, the local vector map, and the goal points, respectively.
We demonstrated the system on a crowded urban road with dense traffic, where \teacher{} controlled the vehicle in closed loop.
\textbf{A demonstration video is available on our project page.}

\section{Related Work}
\label{sec:related-work}

\paragraph{Closed-Loop Reinforcement Learning for Driving.}
Closed-loop reinforcement learning allows driving policies to learn from
states induced by their own actions rather than only from fixed expert
trajectories. Urban Driver~\cite{scheel2021urbandriver} trains policies in
log-initialized urban scenes using policy gradients, while
BC-SAC~\cite{lu2023bcsac} combines imitation with RL on real logs to
improve robustness in safety-critical scenarios. Later work scales
model-free training: CarPlanner~\cite{zhang2025carplanner} makes
large-scale RL tractable through consistent auto-regressive rollouts,
CaRL~\cite{jaeger2025carl} shows that scalable simulation and simple
rewards suffice for effective policy optimization, and
PlannerRFT~\cite{li2026plannerrft} fine-tunes an
imitation-pretrained diffusion planner by efficient closed-loop reinforcement learning. To reduce
interaction cost, Think2Drive~\cite{li2024think2drive},
AdaWM~\cite{wang2025adawm}, and Raw2Drive~\cite{yang2025raw2drive} learn
and explore within world models, extending model-based RL from latent
planning to end-to-end driving. 
Finally, multi-agent training~\cite{zhang2024marlsurvey} replaces
fixed background traffic with learning agents: self-play at scale yields
reliable sim agents~\cite{cornelisse2025simagents} and robust,
naturalistic driving without human data~\cite{cusumano2025selfplay},
while SPICED~\cite{cornelisse2026human} shows that a small amount of
human driving can regularize self-play toward human-compatible
conventions.

\paragraph{Visual Foundation Models for Driving.}
Camera-based perception and end-to-end driving obtain visual knowledge from both general-purpose and driving-specific pretraining. Many systems inherit backbones trained with ImageNet supervision, self-supervised DINO objectives, or vision--language alignment~\citep{russakovsky2015imagenet,caron2021dino,oquab2023dinov2,simeoni2025dinov3,zhai2023siglip,tschannen2025siglip2}. Driving-specific pretraining has further explored self-supervised objectives tailored to driving sensor data~\citep{yang2023bevcontrast,yang2023unipad,yang2024vidar}, as well as geometry-oriented pretraining that injects 3D structure into visual representations~\citep{park2021dd3d,xu2026geometry}.
While these methods design specialized pretraining objectives for driving, a complementary direction is to reuse and consolidate knowledge already encoded by existing foundation models. RADIO consolidates representations from multiple frozen visual foundation models into a single general-purpose backbone through agglomerative distillation, with PHI-S providing feature standardization across heterogeneous teachers~\citep{ranzinger2024amradio,ranzinger2024phis}. Following this recipe, we develop DriveVFM, a visual foundation model tailored to autonomous driving. DriveVFM distills DINOv3, SigLIP2, SAM, and Depth Anything into a shared backbone using their features as supervision~\citep{simeoni2025dinov3,tschannen2025siglip2,kirillov2023sam,yang2024depthanythingv2}. This removes the need for task-specific perception labels and allows pretraining to freely combine web images with driving scenes.

\paragraph{End-to-End Autonomous Driving.}
Most camera-based end-to-end driving policies are trained through behavioral cloning, spanning conditional control, unified perception--planning architectures, and multimodal trajectory generation and selection~\citep{li2025gtrs,codevilla2018end,hu2023uniad,jiang2023vad,jia2025drivetransformer,li2025hydramdp,liao2025diffusiondrive, gao2026takevla, wang2026beyond}. Although these methods differ in how trajectories are represented and decoded, their behavioral supervision largely remains the ego trajectory recorded in human driving logs.
Recent work instead transfers behavior learned through interactive simulation to visual policies. ROACH distills a CARLA RL coach into a monocular policy~\citep{zhang2021roach}; TerraTransfer aligns a visual encoder with a vectorized self-play policy without logged-trajectory targets~\citep{xiong2026terratransfer}; and GigaPixel trains a pixel-based student through self-play DAgger in a simplified renderer~\citep{rowe2026gigapixel}. Pictura further enables self-play directly from perspective images through high-throughput rendering~\citep{yin2026pictura}. Other approaches place visual policies inside photorealistic reconstructed worlds for reinforcement learning, synthetic-data training, or post-training~\citep{gao2025rad, ni2025recondreamerrl, tian2026simscale,li2026worldengine}. \student follows the privileged-to-visual transfer paradigm and independently pretrains the action model and visual representation in their respective learning regimes. \teacher learns goal-conditioned behavior through closed-loop interaction in structured, log-grounded worlds, while DriveVFM learns visual representations from real images. The two are reunited through trajectory distillation, without placing the camera student inside a rendered closed-loop training environment.
\section{Conclusion}
\label{sec:discussion-conclusion}

This report presented an end-to-end driving system whose behavior is not bounded by human demonstrations.
Driving is decomposed into a perception model and an action model, each pretrained in the regime that suits it: DriveVFM learns visual representations from raw images by distilling frozen vision foundation models, and \teacher learns driving behavior from scratch through closed-loop reinforcement learning in log-initialized interactive worlds.
\student reunites the two by distilling the frozen teacher's rollouts into a camera-only planner. On nuPlan, \teacher exceeds the Log-Replay expert on all three community splits in both non-reactive and reactive modes, showing that closed-loop RL can learn behavior beyond the logs that seed its training worlds.
\student achieves state-of-the-art performance on NAVSIMv1, NAVSIMv2, and the closed-loop HUGSIM benchmark without any human trajectory supervision.
We believe this recipe, pretraining perception and action separately and unifying them through distillation, offers a scalable path toward end-to-end driving beyond human demonstrations.
\newpage
\section{Contributors}
\label{sec:contributors}

\newcommand{\contrib}[2]{\noindent\textbf{#1:} #2\par\vspace{0.6em}}

\contrib{DriveRL}{Hao He, Chengcheng Hu, Zirun Su, Heng Zhang}

\contrib{DriveVFM}{Haisong Liu}

\contrib{DriveZero}{Haisong Liu$^*$, Jinke Li$^*$, Haochen Tian$^*$, Zhenwei Shen, Hongyang Li}

\contrib{Real-World Deployment}{Hao He, Zhichao Li, Yunchen Yang, Bochao Huang, Siyu Zhang, Kuangye Chen, Heng Zhang, Xiongjie Zhang, Wentao Dai, Hengchen Dai, Siyuan Liu}

\noindent\textbf{Project Lead:} Hao He, Zhichao Li, Zehao Huang, Naiyan Wang

\noindent\rule{0.3\linewidth}{0.4pt}\\[0.3em]
{\footnotesize $^*$Equal contribution.}

\bibliographystyle{plainnat}
\bibliography{bib_short,main}

\clearpage
\appendix
\setcounter{table}{0}
\renewcommand{\thetable}{A\arabic{table}}
\setcounter{figure}{0}
\renewcommand{\thefigure}{A\arabic{figure}}
\section{\teacher Supplementary Material}
\label{app:teacher-implementation}

\setcounter{table}{0}
\renewcommand{\thetable}{A\arabic{table}}
\makeatletter
\@ifundefined{theHtable}{}{\renewcommand{\theHtable}{A\arabic{table}}}
\makeatother


\subsection{Simulation and Training Infrastructure}
\label{app:simulation-training}

\subsubsection{Distributed PPO Configuration}
\label{app:training-config}

Table~\ref{tab:appendix-training-config} lists the data, rollout, and PPO settings of the DriveRL teacher. Each rank rolls out 2,048 worlds for 110 steps, giving 225,280 transitions per update, which are split into 11 minibatches of 20,480 for each of the four PPO epochs.

\begin{table}[H]
  \caption{\textbf{Training and rollout configuration.} Runtime and optimization settings for \teacher teacher.}
  \label{tab:appendix-training-config}
  \centering
  \footnotesize
  \setlength{\tabcolsep}{4pt}
  \begin{tabularx}{\textwidth}{@{}p{0.22\textwidth}p{0.23\textwidth}p{0.22\textwidth}X@{}}
    \toprule
    \textbf{Setting} & \textbf{Value} & \textbf{Setting} & \textbf{Value} \\
    \midrule
    World initialization & 11 frames / 2~s & Rollout horizon & 110 steps at 5~Hz \\
    Worlds per rank & 2,048 & Traffic quota / rank & 1,024 reactive + 1,024 replay \\
    Distributed scale & 12 nodes / 96 GPUs & Global rollout batch & 196,608 worlds \\
    Nominal transitions / rank & 225,280 & Optimizer minibatch / rank & 20,480 \\
    Minibatches / epoch & 11 & Optimization epochs & 4 \\
    Optimizer steps / update & 44 & Training updates & 2,400 \\
    Training time & approximately 21~h & Optimizer & Muon \\
    Discount $\gamma$ & 0.99 & GAE $\lambda$ & 0.95 \\
    Policy / value clip & 0.2 / 0.2 & Entropy coefficient & 0.01 \\
    Value coefficient & 0.5 & Gradient clipping & 0.5 \\
    Advantage normalization & Valid current-rank minibatch & Return normalization & disabled \\
    Initial learning rate & $3\times10^{-4}$ & LR schedule & cosine, no warmup \\
    \bottomrule
  \end{tabularx}
\end{table}

\subsubsection{Mixed-Agent Environment and Traffic Providers}
\label{app:behavior_provider}

The mixed-agent simulator assigns background actors to behavior providers through a common physical state--action interface. The reported training setup uses log replay, IDM, and front-vehicle braking; the same interface also supports policy-controlled actors for self-play.

\paragraph{IDM.}
The IDM implementation uses an approximately 100~m lane-center search range, a 5~m map-matching threshold, a 1~m minimum gap, a 1.5~s headway, and 0.1~s integration.

\paragraph{Front-vehicle braking.}
To expose the ego to sudden deceleration of a lead vehicle, which is rare in the logs, the simulator can override the behavior of a background vehicle with an emergency-braking provider. At world initialization, each vehicle already present in the scene and located within a 3~m lateral corridor ahead of the ego is marked as a braking candidate with probability 3\%. At every 5~Hz step, each candidate triggers braking with probability 5\%. Once triggered, the vehicle decelerates at a rate capped at $-10\,\mathrm{m/s^2}$ until it comes to a stop, after which it returns to its default provider.


\subsection{Structured Observation and Goal Conditioning}
\label{app:goals-features}

The source loader stores at most 128 actors. The policy filters actors within a 200~m radius, orders them by distance, and retains at most 96 agent tokens with ego fixed in slot 0; 32 slots each are reserved for vehicles and pedestrians. Each actor history contains five frames total, ordered from $t-0.8$~s to $t$ at 0.2~s intervals. The local map window covers 100~m behind, 200~m ahead, and $\pm 200$~m laterally. Table~\ref{tab:appendix-observation-model} summarizes the resulting structured Teacher inputs.

\begin{table}[H]
  \caption{\textbf{Structured Teacher input schema.} Listed capacities are maximum counts after policy filtering.}
  \label{tab:appendix-observation-model}
  \centering
  \footnotesize
  \setlength{\tabcolsep}{3.5pt}
  \begin{tabularx}{\textwidth}{@{}p{0.18\textwidth}p{0.17\textwidth}X p{0.15\textwidth}@{}}
    \toprule
    \textbf{Input} & \textbf{Per-item shape} & \textbf{History or capacity} & \textbf{Frame} \\
    \midrule
    Goal anchors & 2-D point & 2 anchors per policy input & Ego \\
    Ego state & 9-D & Current frame & Ego \\
    Actor history & 15-D / frame & 5 frames; $\leq96$ actors including ego & Ego \\
    Vector-map segment & 10-D / token & $\leq256$ tokens, including $\leq64$ lane centers & Ego \\
    Traffic-light state & 4-D / segment & Attached to the corresponding map segment & Map segment \\
    \bottomrule
  \end{tabularx}
\end{table}

For one policy observation, the goal tensor contains two 2-D points, i.e., $G_t\in\mathbb{R}^{2\times2}$; batch and actor dimensions are added only for storage and export. The map loader exposes at most 1,024 lane-center candidates and 1,024 other segment candidates. It first retains at most 64 lane centers, merges them with the remaining segments, and then applies the common spatial and distance filter to obtain at most 256 map tokens. Each raw map token contains two endpoints (4-D), type (1-D), speed limit (1-D), and traffic-light state (4-D). Segments without an associated signal use a zero traffic-light vector. An MLP with endpoint positional encoding maps each 10-D token to a 256-D embedding. Route intent is extracted into goal anchors without a separate route token.

Goals condition the policy. Their construction differs between training and deployment, as summarized in Table~\ref{tab:appendix-goal-protocol}. Although the goal sources and update schedules differ, both protocols use the same two-anchor, permutation-invariant interface; Section~\ref{app:goal-layout} studies sensitivity to the resulting goal layout.

\begin{table}[H]
  \caption{\textbf{Training and deployment goal protocols.} Goal sources, update schedules, and slot semantics used by the Teacher.}
  \label{tab:appendix-goal-protocol}
  \centering
  \footnotesize
  \setlength{\tabcolsep}{4pt}
  \begin{tabularx}{\textwidth}{@{}p{0.19\textwidth}p{0.36\textwidth}X@{}}
    \toprule
    \textbf{Aspect} & \textbf{PPO training} & \textbf{Formal nuPlan deployment} \\
    \midrule
    Source & Future-log positions & Current ego state + route polyline \\
    Update & Once at reset; fixed per episode & Every planner step \\
    Slots & One or two; duplicate a single goal & Two; near and far route goals \\
    Snapping & 50\% logged / 50\% lane projection & Causal route polyline \\
    Order semantics & Shared encoder + mean pool & Same unordered-set semantics \\
    \bottomrule
  \end{tabularx}
\end{table}

Training candidates are sampled without replacement with weight $p(\tau)\propto(\tau+1)$ and then sorted by time. When lane snapping is selected, the sampler considers the four nearest lane-center projections within 20~m and assigns weight $[\max(1-d/20,0)]^2$; it keeps the original goal if no valid lane is available. At deployment, ego is projected onto the route polyline and the look-ahead interval is
\[
L=\min\left(L_{\mathrm{remain}},12\max(\lVert v_t\rVert,5)\right).
\]
The two goals lie at its midpoint and endpoint. Exchanging the goal slots does not change their mean-pooled embedding, so they represent an unordered spatial set rather than a waypoint sequence.

\subsection{Policy, Action Space, and Vehicle Dynamics}
\label{app:policy-dynamics}

The action head outputs $[\alpha_j,\alpha_{\dot{\delta}},\beta_j,\beta_{\dot{\delta}}]$ and applies $\operatorname{softplus}(x)+1$, ensuring $\alpha,\beta>1$. The normalized jerk and steering-rate dimensions are conditionally independent given the observation and goal. PPO rollouts sample from both Beta distributions; formal nuPlan evaluation executes their per-dimension analytic mode $(\alpha-1)/(\alpha+\beta-2)$. Table~\ref{tab:appendix-action-dynamics} lists the physical interface and dynamics constants.

\begin{table}[H]
  \caption{\textbf{Action and vehicle-dynamics parameters.} Physical command ranges and constants used by the kinematic bicycle model.}
  \label{tab:appendix-action-dynamics}
  \centering
  \footnotesize
  \begin{tabularx}{0.82\textwidth}{@{}Xr l@{}}
    \toprule
    \textbf{Parameter} & \textbf{Value / range} & \textbf{Unit} \\
    \midrule
    Longitudinal jerk & $[-8.00,5.00]$ & $\mathrm{m/s^3}$ \\
    Tire steering rate & $[-0.80,0.80]$ & $\mathrm{rad/s}$ \\
    Positive jerk cap ($a\geq0$) & 1.00 & $\mathrm{m/s^3}$ \\
    Wheelbase & 3.09 & m \\
    Rear axle to center & 1.46 & m \\
    Maximum acceleration & 4.00 & $\mathrm{m/s^2}$ \\
    Maximum steering angle & $\pi/3$ & rad \\
    Maximum steering rate & 0.80 & $\mathrm{rad/s}$ \\
    Acceleration time constant & 0.20 & s \\
    Steering time constant & 0.05 & s \\
    \bottomrule
  \end{tabularx}
\end{table}

An affine transform maps normalized actions into the ranges above. The teacher can export $\alpha$, $\beta$, mean, mode, and sampled actions for student distillation. Exporting both concentration parameters preserves more information about the policy distribution than exporting only the mean, mode, or sampled action.


\subsection{Reward, Value Decomposition, and PPO Objective}
\label{app:reward-ppo}

The reported teacher uses the scalar reward
\begin{equation}
  r_t
  =
  h_t
  +(1-d_t)
  \left(
    g_t+\frac{1}{110}\prod_{k\in\mathcal{K}}q_{t,k}
  \right),
  \qquad
  \mathcal{K}
  =
  \{\text{cross-lane},\text{centerline},\text{curb},
  \text{comfort},\text{TTC},\text{overspeed}\}.
  \label{eq:appendix-teacher-reward}
\end{equation}

Here, $d_t$ is one when a hard event terminates the rollout. The reward terms are configured as follows:
\begin{itemize}
  \item \textbf{Hard:} newly occurring polygon collisions receive a penalty of $-1.0$; off-road events terminate the rollout and use a speed-dependent penalty with base value $-2.0$.
  \item \textbf{Goal:} first arrival within $1.5$~m yields a one-time reward of $0.5$. Goal-distance shaping and survival rewards are disabled, and goal arrival does not terminate the rollout.
  \item \textbf{Soft:} each score $q_{t,k}\in[0,1]$ is produced by its corresponding driving-quality calculator. Their product is normalized by the 110-step rollout horizon.
\end{itemize}

The multiplicative soft term requires the six driving-quality criteria to be jointly satisfactory, preventing strong performance on one criterion from compensating for a serious deficiency in another. The resulting scalar return, however, combines sparse hard events, sparse goal arrivals, and a dense joint soft signal in a single regression target. We therefore use an eight-channel critic,
\begin{equation}
  \mathcal{C}
  =
  \{\mathrm{hard},\mathrm{goal}\}\cup\mathcal{K},
  \qquad
  \mathbf{V}(s_t)
  =
  [V_c(s_t)]_{c\in\mathcal{C}}
  \in\mathbb{R}^{8}.
  \label{eq:appendix-value-channels}
\end{equation}

The hard and goal channels retain their reward semantics. The six soft channels do not correspond to separately defined rewards. They partition the single multiplicative soft reward according to a fixed deficit-based rule, solely to provide structured critic targets while preserving the scalar reward. Specifically, let
\begin{equation}
\begin{aligned}
  f_t
  &=
  (1-d_t)\frac{1}{110}
  \prod_{j\in\mathcal{K}}q_{t,j},
  &
  b_{t,k}
  &=
  \max(1-q_{t,k},0),
  &
  B_t
  &=
  \sum_{j\in\mathcal{K}}b_{t,j},
  \\
  w_{t,k}
  &=
  \begin{cases}
    \dfrac{b_{t,k}}{B_t}, & B_t>\varepsilon_b,\\[5pt]
    \dfrac{1}{|\mathcal{K}|}, & B_t\leq\varepsilon_b,
  \end{cases}
  &
  r_{t,c}
  &=
  \begin{cases}
    h_t, & c=\mathrm{hard},\\
    (1-d_t)g_t, & c=\mathrm{goal},\\
    w_{t,k}f_t, & c=k\in\mathcal{K}.
  \end{cases}
\end{aligned}
\label{eq:appendix-reward-components}
\end{equation}

Here, $\varepsilon_b=10^{-8}$. The uniform fallback avoids an undefined allocation when all six soft scores are perfect. Since $\sum_{k\in\mathcal{K}}w_{t,k}=1$, the eight component rewards exactly conserve the scalar reward, i.e., $\sum_{c\in\mathcal{C}}r_{t,c}=r_t$.

Generalized advantage estimation is applied channel-wise using the same discount, trace parameter, and terminal boundary:
\begin{align}
  \delta_{t,c}
  &=
  r_{t,c}
  +\gamma n_tV_c(s_{t+1})
  -V_c(s_t),
  \nonumber\\
  A_{t,c}
  &=
  \delta_{t,c}
  +\gamma\lambda n_tA_{t+1,c},
  \qquad
  c\in\mathcal{C},
  \quad
  n_t=1-d_t.
  \label{eq:appendix-component-gae}
\end{align}

The final rollout state supplies the usual bootstrap value. We use $\gamma=0.99$ and $\lambda=0.95$, and train each value channel toward $\widehat{R}_{t,c}=A_{t,c}+V_c(s_t)$.

The actor uses the summed advantage $A_t=\sum_{c\in\mathcal{C}}A_{t,c}$, normalized over valid samples within each minibatch, in the standard clipped PPO surrogate. Because the component rewards sum to $r_t$ and all channels share the same $(\gamma,\lambda,n_t)$, linearity of GAE makes this summed advantage equivalent, up to floating-point error, to scalar GAE applied to $r_t$ and the total value $V_{\mathrm{tot}}(s_t)=\sum_{c\in\mathcal{C}}V_c(s_t)$. For critic optimization, a PPO-style clipped value loss is computed independently for each channel, and the eight losses are summed before applying the value-loss coefficient. The decomposition therefore structures critic supervision and diagnostics while preserving the scalar policy objective in Eq.~\eqref{eq:appendix-teacher-reward}.

Traffic-light violations are reported separately as a diagnostic metric; they are not part of the reported teacher's scalar reward. PPO and distributed settings appear in Table~\ref{tab:appendix-training-config}.


\subsection{Additional DriveRL Analyses}
\label{app:teacher-analyses}

Unless otherwise stated, analyses in this section use the reported DriveRL checkpoint. Some controlled ablations were conducted in an earlier 32-GPU training setup and are used only for within-table comparisons; their absolute scores should not be compared directly with the reported teacher.

\subsubsection{Detailed nuPlan Metric Breakdown}
\label{app:detailed-nuplan}

Table~\ref{tab:appendix-single-ego-results} reports the component metrics for all six nuPlan evaluation settings.

\begin{table}[H]
  \caption{\textbf{Detailed nuPlan metrics for \teacher.} Results across six evaluation settings; higher is better for every column.}
  \label{tab:appendix-single-ego-results}
  \centering
  \footnotesize
  \setlength{\tabcolsep}{2.8pt}
  \begin{tabular*}{\textwidth}{@{\extracolsep{\fill}}lccccccccc@{}}
    \toprule
    Split / mode & Score & Collision & Drivable & Direction & \makecell{Making\\progress} & TTC & Progress & Speed & Comfort \\
    \midrule
    Val14 NR          & \textbf{95.16} & 98.75 & 100.00 & 99.60 & 98.93  & 95.97 & 96.30 & 98.80 & 98.84 \\
    Val14 R           & \textbf{94.25} & 98.57 & 100.00 & 99.24 & 98.75  & 95.80 & 93.83 & 99.21 & 98.66 \\
    Test14-hard NR    & \textbf{89.97} & 98.90 & 98.90  & 98.16 & 98.16  & 93.38 & 91.74 & 97.67 & 95.96 \\
    Test14-hard R     & \textbf{89.18} & 97.43 & 99.26  & 98.53 & 96.69  & 94.49 & 88.73 & 98.69 & 96.32 \\
    Test14-random NR  & \textbf{94.50} & 98.08 & 99.23  & 99.81 & 100.00 & 95.40 & 95.48 & 98.33 & 98.85 \\
    Test14-random R   & \textbf{95.00} & 98.85 & 99.62  & 100.00 & 99.62 & 96.55 & 93.34 & 98.85 & 98.85 \\
    \bottomrule
  \end{tabular*}
\end{table}

\subsubsection{Self-Play Study}
\label{app:selfplay-study}

The \teacher-SelfPlay assigns the same policy to the ego vehicle and up to ten vehicle NPCs. Four IDM-controlled vehicles are first included in the candidate set, and the remaining slots are filled with the most relevant policy-controlled vehicles. Table~\ref{tab:appendix-selfplay} summarizes the configuration differences between \teacher-SelfPlay and \teacher.

\begin{table}[H]
  \caption{\textbf{Configuration differences between \teacher and \teacher-SelfPlay.} The two policies use the same mixed-agent simulator and PPO objective.}
  \label{tab:appendix-selfplay}
  \centering
  \footnotesize
  \setlength{\tabcolsep}{4pt}
  \begin{tabularx}{\textwidth}{@{}p{0.27\textwidth}XX@{}}
    \toprule
    \textbf{Setting} & \textbf{Single-Ego Teacher} & \textbf{Self-Play Teacher} \\
    \midrule
    Policy-controlled actors & Ego only & Ego + $\leq10$ vehicle NPCs \\
    Traffic composition & Log/IDM mixed50 & Policy NPCs + log/IDM mixed50 \\
    Training samples & 922,703 & 909,645 (subset of Single-Ego data) \\
    Worlds per rank & 2,048 & 512 \\
    Global rollout worlds & 196,608 & 49,152 \\
    \bottomrule
  \end{tabularx}
\end{table}

The following comparison reports the Single-Ego baseline and the Self-Play variants. Among the latter, disabling history dropout, dynamics noise, and front-vehicle braking gives the highest mean score. Its mean performance is comparable to that of the Single-Ego baseline, demonstrating the feasibility of the one-policy, per-actor formulation but not a consistent performance advantage from Self-Play.

\begin{table}[H]
  \caption{\textbf{Self-Play Comparison on the Six Closed-Loop Evaluations.} All Self-Play variants are trained for 4,800 updates; the Single-Ego baseline uses its reported 2,400-update checkpoint. The best checkpoint is reported for each run, and Mean is the unweighted average over the six evaluations.}
  \label{tab:self-play-comparison}
  \centering
  \footnotesize
  \resizebox{\textwidth}{!}{%
  \begin{tabular}{lccccccc}
    \toprule
    \multirow{2}{*}[-0.15ex]{\textbf{Model}} & \multicolumn{2}{c}{\textbf{Val14} $\uparrow$} & \multicolumn{2}{c}{\textbf{Test14-hard} $\uparrow$} & \multicolumn{2}{c}{\textbf{Test14-random} $\uparrow$} & \multirow{2}{*}[-0.15ex]{\textbf{Mean} $\uparrow$} \\
    \cmidrule(lr){2-3} \cmidrule(lr){4-5} \cmidrule(lr){6-7}
    & \textbf{NR} & \textbf{R} & \textbf{NR} & \textbf{R} & \textbf{NR} & \textbf{R} & \\
    \midrule
    \teacher & \textbf{95.16} & 94.25 & 89.97 & \textbf{89.18} & 94.50 & 95.00 & 93.01 \\
    Self-Play, 3-perturb + augmentation & 95.08 & \textbf{94.48} & 88.30 & 87.27 & 94.60 & 94.68 & 92.40 \\
    Self-Play, 3-perturb & 95.15 & 93.74 & 88.90 & 87.47 & 93.70 & 94.77 & 92.29 \\
    Self-Play, augmentation & 94.91 & 94.38 & 88.56 & 87.92 & 93.43 & 94.22 & 92.24 \\
    Self-Play & 95.09 & 93.92 & \textbf{90.04} & 88.96 & \textbf{95.04} & \textbf{95.63} & \textbf{93.11} \\
    \bottomrule
  \end{tabular}%
  }
\end{table}

\subsubsection{Traffic-Light Input and Reward}
\label{app:traffic-light-studies}

We study traffic-light conditioning using a separate controlled training setup; absolute scores in this subsection are therefore not directly comparable to those of the reported teacher. We first test the effect of adding traffic-light state to the policy. Each map segment carries a four-dimensional unknown/red/yellow/green feature. Table~\ref{tab:tl-input} compares policies with and without this feature.

\begin{table}[H]
  \caption{\textbf{Traffic-Light Input Ablation.} The red-light detector is diagnostic only in both settings.}
  \label{tab:tl-input}
  \centering
  \footnotesize
  \setlength{\tabcolsep}{3pt}
  \begin{tabular*}{\linewidth}{@{\extracolsep{\fill}}lcccccc@{}}
    \toprule
    \multirow{2}{*}[-0.15ex]{\textbf{TL input}} & \multicolumn{2}{c}{\textbf{Val14} $\uparrow$} & \multicolumn{2}{c}{\textbf{Test14-hard} $\uparrow$} & \multicolumn{2}{c}{\textbf{Test14-random} $\uparrow$} \\
    \cmidrule(lr){2-3} \cmidrule(lr){4-5} \cmidrule(lr){6-7}
    & \textbf{NR} & \textbf{R} & \textbf{NR} & \textbf{R} & \textbf{NR} & \textbf{R} \\
    \midrule
    Off        & 93.35 & 93.19 & 86.49 & 85.41 & 93.48 & 92.24 \\
    On         & 94.76 & 93.62 & 88.35 & 85.70 & 95.21 & 93.89 \\
    Difference & $+1.41$ & $+0.43$ & $+1.86$ & $+0.29$ & $+1.73$ & $+1.65$ \\
    \bottomrule
  \end{tabular*}
\end{table}

Exposing traffic-light state improves scores in all six evaluation settings.

Within the same setup, we next add a red-light training objective to a policy with traffic-light input. The table reports the policy-only red-light violation rate, which counts violations in scenes where the logged human remains compliant, following NAVSIM's human-reference filtering principle~\citep{Dauner2024navsim}.

\begin{table}[H]
  \caption{\textbf{Traffic-Light Reward and Violation-Termination Ablation.} All runs use 32 GPUs; ``pp'' denotes percentage points.}
  \label{tab:tl-reward}
  \centering
  \footnotesize
  \setlength{\tabcolsep}{2pt}
  \begin{tabular*}{\linewidth}{@{\extracolsep{\fill}}lccccccc@{}}
    \toprule
    \multirow{2}{*}[-0.15ex]{\textbf{Objective}} & \multirow{2}{*}[-0.15ex]{\makecell{\textbf{Policy-only}\\\textbf{violation}}} & \multicolumn{2}{c}{\textbf{Val14} $\uparrow$} & \multicolumn{2}{c}{\textbf{Test14-hard} $\uparrow$} & \multicolumn{2}{c}{\textbf{Test14-random} $\uparrow$} \\
    \cmidrule(lr){3-4} \cmidrule(lr){5-6} \cmidrule(lr){7-8}
    & & \textbf{NR} & \textbf{R} & \textbf{NR} & \textbf{R} & \textbf{NR} & \textbf{R} \\
    \midrule
    TL input only        & 3.88\% & 94.29 & 93.19 & 84.85 & 85.65 & 93.79 & 93.09 \\
    + TL reward          & 2.03\% & 93.08 & 91.35 & 81.71 & 82.96 & 92.36 & 91.79 \\
    Difference           & $-1.85$ pp & $-1.21$ & $-1.84$ & $-3.14$ & $-2.69$ & $-1.43$ & $-1.30$ \\
    \bottomrule
  \end{tabular*}
\end{table}

The traffic-light reward lowers the policy-only violation rate from 3.88\% to 2.03\%, while all six aggregate scores decrease. This trade-off shows that the aggregate nuPlan score alone does not capture the improvement in red-light compliance, since traffic-light violations are not part of the benchmark score. We therefore report the violation rate alongside the aggregate metrics.

\subsubsection{Goal-Anchor Layout}
\label{app:goal-layout}

\teacher always retains two goal input slots. A single-anchor condition duplicates the same goal into both slots and therefore leaves the network architecture unchanged. The formal route-goal interface produces a near anchor $g_{\mathrm{near}}$ and a far anchor $g_{\mathrm{far}}$ at every planner step.

\paragraph{Fixed-checkpoint sensitivity.}
We first fix the reported \teacher checkpoint and vary only the goal layout used for nuPlan evaluation. The default uses the near and far anchors, while the other settings duplicate either anchor into both slots.

\begin{table}[H]
  \caption{\textbf{Goal-Anchor Layout Evaluation for the Fixed Single-Ego Teacher.} The highest score in each row is bold.}
  \label{tab:appendix-goal-eval}
  \centering
  \footnotesize
  \setlength{\tabcolsep}{4pt}
  \begin{tabularx}{\textwidth}{@{}Xccc@{}}
    \toprule
    \textbf{Split / mode} & \textbf{Default $[g_{\mathrm{near}},g_{\mathrm{far}}]$} & \textbf{Near $[g_{\mathrm{near}},g_{\mathrm{near}}]$} & \textbf{Far $[g_{\mathrm{far}},g_{\mathrm{far}}]$} \\
    \midrule
    Val14 NR         & \textbf{95.16} & 94.97 & 94.36 \\
    Val14 R          & \textbf{94.25} & 94.00 & 93.22 \\
    Test14-hard NR   & 89.97 & \textbf{90.74} & 88.01 \\
    Test14-hard R    & \textbf{89.18} & 88.82 & 87.13 \\
    Test14-random NR & 94.50 & \textbf{94.57} & 93.18 \\
    Test14-random R  & 95.00 & \textbf{95.23} & 92.84 \\
    \bottomrule
  \end{tabularx}
\end{table}

The default and near-only layouts have similar average performance, while duplicating the far anchor lowers all six scores. This indicates that the near anchor provides important local guidance; the far anchor is most useful when paired with, rather than substituted for, the near anchor.

\paragraph{Training-distribution ablation.}
We additionally study whether this preference is consistent with the goal distribution used during training, using a separate controlled training setup. Absolute scores in this experiment are therefore not directly comparable to those of the reported teacher. The default distribution samples one or two future-log goals with equal probability and is evaluated with the near--far pair. The other models always sample two goals, retain either the earlier or later one, duplicate it into both slots, and are evaluated with the corresponding near--near or far--far layout.

\begin{table}[H]
  \caption{\textbf{Goal-Anchor Training Distribution Ablation.} Models are trained from scratch with 32-GPU Adam, and evaluation layouts are matched to the corresponding training distributions. Mean is the unweighted average over the six evaluations.}
  \label{tab:appendix-goal-training}
  \centering
  \footnotesize
  \setlength{\tabcolsep}{2.5pt}
  \begin{tabular*}{\textwidth}{@{\extracolsep{\fill}}llccccccc@{}}
    \toprule
    \multirow{2}{*}[-0.15ex]{\textbf{Training distribution}} & \multirow{2}{*}[-0.15ex]{\textbf{Evaluation}} & \multicolumn{2}{c}{\textbf{Val14} $\uparrow$} & \multicolumn{2}{c}{\textbf{Test14-hard} $\uparrow$} & \multicolumn{2}{c}{\textbf{Test14-random} $\uparrow$} & \multirow{2}{*}[-0.15ex]{\textbf{Mean} $\uparrow$} \\
    \cmidrule(lr){3-4} \cmidrule(lr){5-6} \cmidrule(lr){7-8}
    & & \textbf{NR} & \textbf{R} & \textbf{NR} & \textbf{R} & \textbf{NR} & \textbf{R} & \\
    \midrule
    Single / dual (1:1, default) & Near + far & \textbf{94.76} & \textbf{93.62} & \textbf{88.35} & 85.70 & \textbf{95.21} & \textbf{93.89} & \textbf{91.92} \\
    Earlier of two duplicated & Near + near & 94.23 & 93.35 & 86.78 & \textbf{87.38} & 94.73 & 93.23 & 91.62 \\
    Later of two duplicated & Far + far & 93.04 & 92.43 & 83.10 & 82.31 & 93.27 & 90.95 & 89.18 \\
    \bottomrule
  \end{tabular*}
\end{table}

The mixed distribution achieves the highest mean score. With evaluation matched to training, earlier-goal duplication remains close to the default, whereas later-goal duplication performs substantially worse.

\section{\student Supplementary Material}

\subsection{Models and Training}

Table~\ref{tab:appendix-student-implementation} summarizes the model and training hyperparameters of the camera-only \student-Scale{}.

\begin{table}[H]
  \centering
  \caption{\textbf{Model and Training Configuration for \student-Scale{}.}}
  \label{tab:appendix-student-implementation}
  \footnotesize
  \setlength{\tabcolsep}{4pt}
  \begin{tabularx}{0.92\textwidth}{@{}p{0.30\textwidth}X@{}}
    \toprule
    \textbf{Hyperparameter} & \textbf{Value} \\
    \midrule
    \multicolumn{2}{@{}l}{\textbf{Model Configuration}} \\
    Sensors & \cmtt{CAM\_F0}, \cmtt{CAM\_B0}, \cmtt{CAM\_L0}, \cmtt{CAM\_R0} \\
    Input resolution & $960\times512$ (W$\times$H) \\
    Prediction horizon / frequency & 20 steps at $5\,\mathrm{Hz}$ \\
    Trajectory proposals & 64 \\
    Planning representation & 256 dimensions \\
    Proposal decoder & 4 layers \\
    Register tokens & 16 per camera \\
    Visual backbone & DriveVFM ViT-L (frozen) \\
    Backbone adaptation & Q/V LoRA, rank 32 \\
    Trainable parameters & $18.58\,\mathrm{M}$ \\
    Full parameters & $338.46\,\mathrm{M}$ \\
    \midrule
    \multicolumn{2}{@{}l}{\textbf{Training Configuration}} \\
    Training data & 100K \cmtt{navtrain} + 237K \cmtt{SimScale} scenes \\
    GPUs & $16 \times \text{H20}$ \\
    Epochs & 25 \\
    Batch size & 256 \\
    Optimizer & AdamW \\
    Initial learning rate & $2\times10^{-4}$ \\
    Learning-rate schedule & Cosine annealing to zero (no warmup) \\
    Weight decay & $0.01$ \\
    Gradient clipping & Global norm $1.0$ \\
    Training precision & FP32 \\
    Training time & 38h\\
    Trajectory supervision & Teacher rollouts; no human trajectories \\
    \bottomrule
  \end{tabularx}
\end{table}

\subsection{Visual Backbone Scaling}
\label{app:student-backbone-scaling}

We study visual-backbone scaling by comparing DriveVFM ViT-S, ViT-B, and ViT-L. All three backbones are pretrained with the same data sources and identical mixture ratios. For downstream training, all three \student{} variants use only the NAVSIM \cmtt{navtrain} split, without any SimScale data. The pretrained backbone weights remain frozen, and the same rank-32 Q/V LoRA adapters are fine-tuned for every backbone. All other settings are held fixed. Table~\ref{tab:student-backbone-scaling} reports the 6 PDM components and the aggregate PDMS.

\begin{table}[H]
  \caption{Visual-backbone scaling for \student{} on the NAVSIMv1 navtest benchmark. The backbones use the same data sources and identical mixture ratios. All downstream \student{} variants use NAVSIM \cmtt{navtrain} without SimScale and fine-tune rank-32 Q/V LoRA adapters.}
  \label{tab:student-backbone-scaling}
  \centering
  \footnotesize
  \setlength{\tabcolsep}{5pt}
  \begin{tabular*}{\textwidth}{@{\extracolsep{\fill}}c|cccccc|c@{}}
    \toprule
    Backbone & NC $\uparrow$ & DAC $\uparrow$ & TTC $\uparrow$ & Comfort $\uparrow$ & EP $\uparrow$ & DDC $\uparrow$ & PDMS $\uparrow$ \\
    \midrule
    DINOv3 ViT-L & 98.80 & \textbf{99.25} & 95.74 & 99.94 & 92.72 & 95.87 & 94.55 \\
    \midrule
    DriveVFM ViT-S & \textbf{99.17} & 99.14 & \textbf{96.29} & 99.98 & 91.79 & \textbf{95.97} & 94.41 \\
    DriveVFM ViT-B & 98.89 & 99.24 & 95.93 & 99.97 & 92.44 & 95.79 & 94.54 \\
    DriveVFM ViT-L & 99.00 & 99.24 & 95.97 & \textbf{99.99} & \textbf{93.10} & 95.89 & \textbf{94.83} \\
    \bottomrule
  \end{tabular*}
\end{table}

PDMS increases from 94.41 with ViT-S to 94.54 with ViT-B and 94.83 with ViT-L. The 0.42-point gain from ViT-S to ViT-L shows a positive scaling trend across these backbones. Furthermore, this scaling trend holds even against stronger general-purpose backbones: our largest ViT-L variant still edges out DINOv3 ViT-L (94.55 PDMS) despite the latter's much larger pretraining scale.

\subsection{QK-Clip for Stable Training and FP16 Deployment}
\label{app:qk-clip}

Training transformers can cause query--key dot products to grow in some attention heads. The resulting large pre-softmax logits lead to loss spikes and make optimization sensitive to numerical precision. Query--Key Normalization~\citep{henry2020query,dehghani2023scaling} addresses this issue by changing the attention computation itself. DriveVFM instead uses QK-Clip, inspired by Kimi K2~\citep{moonshotai2025kimi}: after an optimizer update, it monitors the maximum pre-softmax logit in each head and rescales the query and key projection weights when the value exceeds a target threshold of 1,000. The query and key receive the same square-root scaling factor, while the value projection and the forward attention formulation remain unchanged. This provides a direct bound on the effective QK scale without introducing an additional normalization operation.

\begin{figure}[H]
  \noindent
  \begin{minipage}[t]{0.35\textwidth}
    \vspace{0pt}
    \centering
    \includegraphics[width=\linewidth]{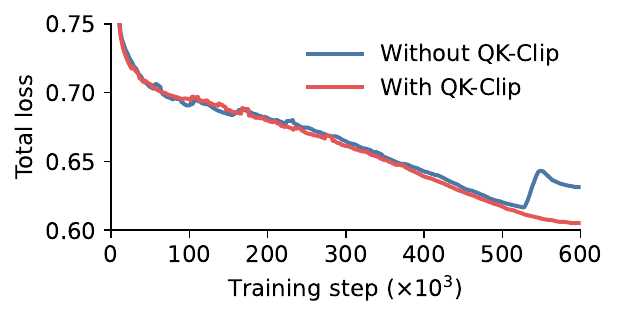}
  \end{minipage}\hfill
  \begin{minipage}[t]{0.55\textwidth}
    \vspace{5pt}
    \centering
    \footnotesize
    \setlength{\tabcolsep}{5pt}
    \renewcommand{\arraystretch}{1.12}
    \begin{tabular}{l|r|rrr}
      \toprule
      Method & \makecell{Max QK Logit} & \makecell{Max Abs} & \makecell{Mean Abs} & Cosine \\
      \midrule
      Baseline & 308{,}320 & 1.243 & 0.109 & 0.893 \\
      QK-Norm & 5{,}125 & 0.768 & 0.032 & 0.981 \\
      QK-Clip & \textbf{995} & \textbf{0.047} & \textbf{0.004} & \textbf{0.999} \\
      \bottomrule
    \end{tabular}
  \end{minipage}

  \noindent
  \begin{minipage}[t]{0.35\textwidth}
    \captionof{figure}{Training loss with and without QK-Clip. QK-Clip stabilizes training, avoiding the loss spikes seen without it.}
    \label{fig:appendix-qk-clip-stability}
  \end{minipage}\hfill
  \begin{minipage}[t]{0.55\textwidth}
    \captionof{table}{FP32-to-FP16 feature deviation. QK-Clip yields the lowest deviation, showing that bounding logits also improves FP16 robustness.}
    \label{tab:appendix-qk-clip}
  \end{minipage}
\end{figure}

Figure~\ref{fig:appendix-qk-clip-stability} compares optimization with and without QK-Clip. Constraining the query--key scale produces a smoother descent and a lower loss in the latter part of training, supporting its role in stabilizing the training.

The same bound is important at deployment. Converting the backbone from FP32 to FP16 can perturb attention logits and propagate errors through the feature stack. We therefore compare FP32 and FP16 backbone outputs under three training configurations: baseline, with QK normalization, and with QK-Clip. Table~\ref{tab:appendix-qk-clip} reports the maximum and mean absolute deviations together with cosine similarity. The reported maximum QK logit is the largest pre-softmax query--key dot product observed across all attention layers. QK-Norm lowers the maximum QK logit to 5,125 and improves both error metrics (mean abs 0.032, cosine 0.981), but a substantial residual error remains, showing it does not fully resolve the instability. QK-Clip further reduces the mean absolute error to $0.004$ and preserves a cosine similarity of $0.999$, indicating near-lossless FP16 deployment.

\subsection{More Visualizations of \student}

\label{app:student-visualization}

\begin{figure}[H]
  \centering
  \includegraphics[width=1\textwidth]{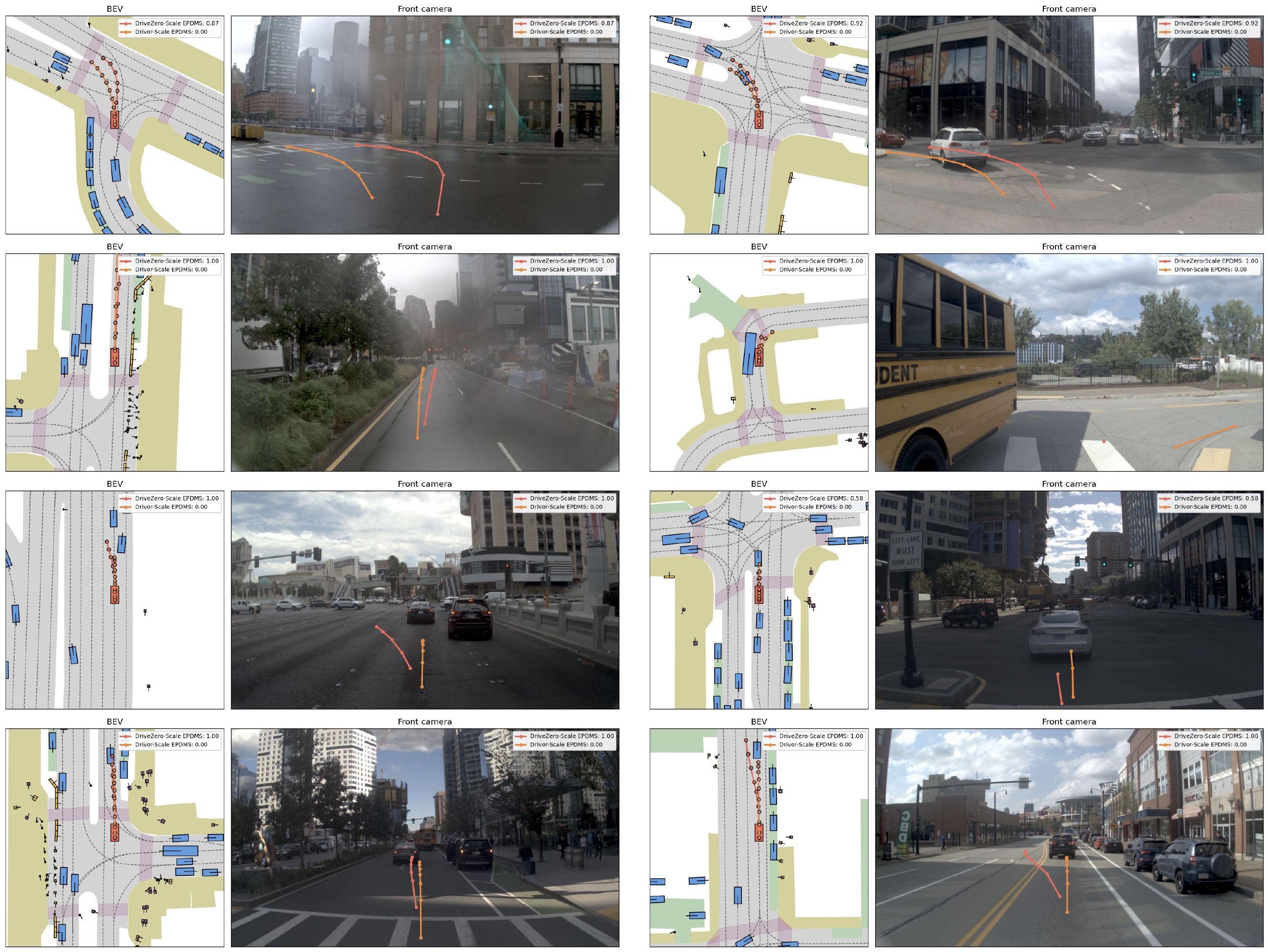}
  \caption{\textbf{More Visualization Results of \textcolor{modelclr}{\student-Scale} vs. \textcolor{drivorclr}{Previous SOTA Model} in NAVSIM navtest~\cite{Dauner2024navsim} Benchmark.}}
  \label{fig:app-student-sota-navsim-visualization}
\end{figure}

\end{document}